\documentclass[10pt,journal,compsoc]{IEEEtran}
\ifCLASSOPTIONcompsoc
  \usepackage[nocompress]{cite}
\else
  \usepackage{cite}
\fi
\ifCLASSINFOpdf
\else
\fi

\usepackage{amsmath,amsfonts}
\usepackage{algorithmic}
\usepackage{algorithm}
\usepackage{array}
\usepackage[caption=false,font=normalsize,labelfont=sf,textfont=sf]{subfig}
\usepackage{textcomp}
\usepackage{stfloats}
\usepackage{url}
\usepackage{verbatim}
\usepackage{graphicx}
\usepackage{cite}

\usepackage{pifont} 
\usepackage{bm}
\usepackage{color}
\usepackage{multirow}
\usepackage{multicol}

\usepackage{booktabs}
\usepackage{epsfig}
\usepackage{multirow}
\usepackage{multicol}
\usepackage{enumerate}
\usepackage{array}
\usepackage[pagebackref,breaklinks,colorlinks]{hyperref}
\usepackage[capitalize]{cleveref}

\usepackage[capitalize]{cleveref}
\crefname{section}{Sec.}{Secs.}
\Crefname{section}{Section}{Sections}
\Crefname{table}{Table}{Tables}
\crefname{table}{Tab.}{Tabs.}

\usepackage{soul,color,xcolor}      
\definecolor{myColor}{rgb}{0,0,0}        
\makeatletter
\newcommand*{\revised@}{\@ifnextchar\bgroup{\revised}{\color{myColor}}}
\newcommand{\revised}[1]{\textcolor{myColor}{#1}}
\makeatother

\begin{document}
%

\title{Learning Efficient Meshflow and Optical Flow from Event Cameras}  %

\author{Xinglong~Luo,
        Ao~Luo,
        Kunming~Luo,
        Zhengning~Wang,~\IEEEmembership{Senior Member,~IEEE,}
        Ping~Tan,~\IEEEmembership{Senior Member,~IEEE,}
        Bing~Zeng,~\IEEEmembership{Fellow,~IEEE,}
        and~Shuaicheng~Liu,~\IEEEmembership{Senior Member,~IEEE}

\IEEEcompsocitemizethanks{

\IEEEcompsocthanksitem Xinglong Luo, Zhengning Wang, Bing Zeng and Shuaicheng Liu are with Institute of Image Processing, School of Information and Communication Engineering, University of Electronic Science and Technology of China, 611731, Chengdu, Sichuan, China. Zhengning Wang is also with Yibin Institute of UESTC, Yibin, 644000, Sichuan, China.

\IEEEcompsocthanksitem Ao Luo is with School of Computing and Artificial Intelligence, Southwest Jiaotong University, 610031, Chengdu, Sichuan, China.

\IEEEcompsocthanksitem Kunming Luo and Ping Tan are with Department of Electronic and Computer Engineering, The Hong Kong University of Science and Technology, 999077, Hongkong, China.

\IEEEcompsocthanksitem This work was supported by National Natural Science Foundation of China (NSFC) under grant Nos. 62372091 and 62402402, Sichuan Science and Technology Program of China under grants Nos. 2023ZDZX0022 and 2022YFQ0079, in part by Hainan Province Key R\&D Program
under grant No. ZDYF2024(LALH)001.

\IEEEcompsocthanksitem Corresponding Authors: Shuaicheng Liu (liushuaicheng@uestc.edu.cn),  Zhengning Wang (zhengning.wang@uestc.edu.cn).
}
}
\IEEEtitleabstractindextext{%
\begin{abstract}

In this paper, we explore the problem of event-based meshflow estimation, a novel task that involves predicting a spatially smooth sparse motion field from event cameras. To start, we review the state-of-the-art in event-based flow estimation, highlighting two key areas for further research: {\bf i)} the lack of meshflow-specific event datasets and methods, and {\bf ii)} the underexplored challenge of event data density. First, we generate a large-scale High-Resolution Event Meshflow (HREM) dataset, which showcases its superiority by encompassing the merits of high resolution at 1280$\times$720, handling dynamic objects and complex motion patterns, and offering both optical flow and meshflow labels. These aspects have not been fully explored in previous works. Besides, we propose Efficient Event-based MeshFlow (EEMFlow) network, a lightweight model featuring a specially crafted encoder-decoder architecture to facilitate swift and accurate meshflow estimation. Furthermore, we upgrade EEMFlow network to support dense event optical flow, in which a Confidence-induced Detail Completion (CDC) module is proposed to preserve sharp motion boundaries. We conduct comprehensive experiments to show the exceptional performance and runtime efficiency (30$\times$ faster) of our EEMFlow model compared to the recent state-of-the-art flow method. As an extension, we expand HREM into HREM+, a multi-density event dataset contributing to a thorough study of the robustness of existing methods across data with varying densities, and propose an Adaptive Density Module (ADM) to adjust the density of input event data to a more optimal range, enhancing the model's generalization ability. We empirically demonstrate that ADM helps to significantly improve the performance of EEMFlow and EEMFlow+ by 8\% and 10\%, respectively. Code and dataset are released at \url{https://github.com/boomluo02/EEMFlowPlus}.
\end{abstract}

}

\maketitle

\IEEEdisplaynontitleabstractindextext

%
\IEEEpeerreviewmaketitle

\IEEEraisesectionheading{\section{Introduction}\label{sec:introduction}}

%
%
%
%
\IEEEPARstart{M}{eshflow}, a spatially smooth sparse motion field, represents motion vectors exclusively at mesh vertices~\cite{liu2016meshflow,Luo_2024_EEMFlow}, which has been widely applied in various vision applications, such as image alignment~\cite{liu2022content,nie2024semi}, video stabilization~\cite{wang2022stable,liu2024minimum} and high dynamic range (HDR) imaging~\cite{yan2017hdr,liu2022ghost}. This motion representation combines the benefits of optical flow and global homography, effectively reducing redundancy in motion information and computational costs, while also accommodating non-rigid motions beyond single-plane movements. However, meshflow estimation on RGB images often encounter challenges under scenarios such as low-light and rapid motions due to the loss of fine image texture details and motion blurs.


\begin{figure}
\centering
\includegraphics[width=\linewidth]{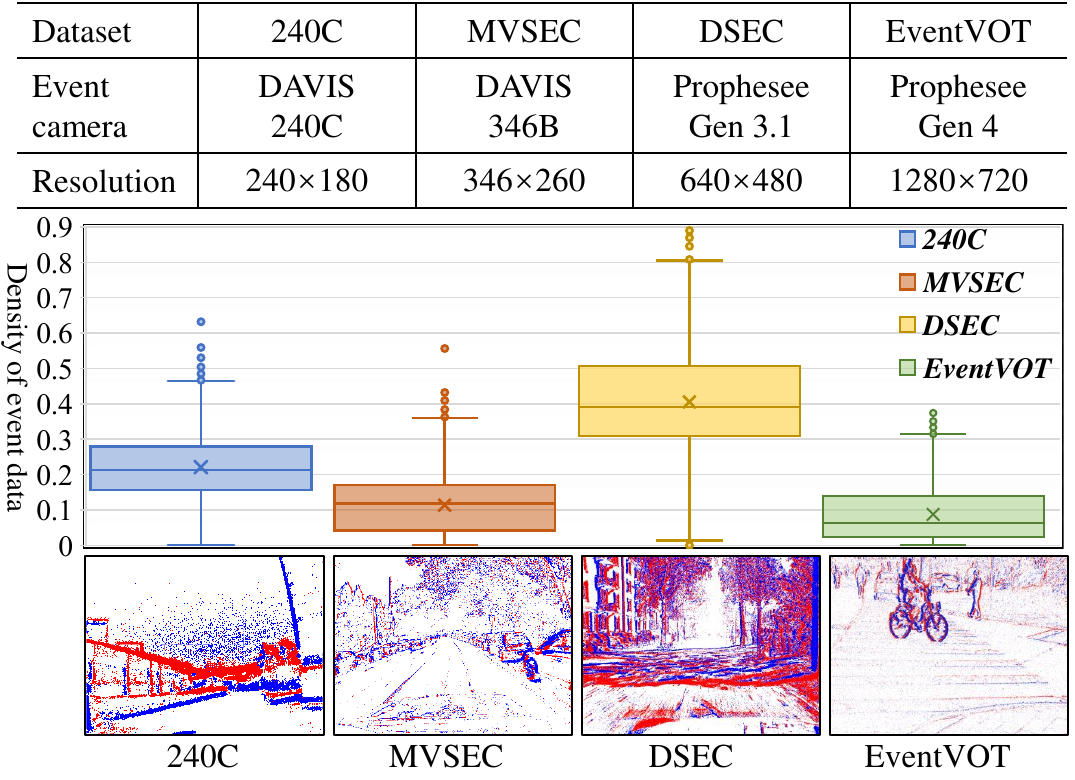}
\caption{\revised{The line chart about the event data density for four real-world datasets, including 240C~\cite{mueggler2017event}, MVSEC~\cite{zhu2018multivehicle}, DSEC~\cite{gehrig2021dsec} and EventVOT~\cite{wang2024event}. These datasets are captured using four different event cameras, and the density ranges of the event data show little overlap and significant differences, while the resolutions also vary.}}
\label{fig:teaser}
\end{figure}

In contrast, event cameras~\cite{gehrig2024low} are well-suited for motion estimation under such situations~\cite{gallego2020eventsurvey, gehrig2021raft}. Equipped with bio-inspired vision sensors, event cameras can generate sequences of events with microsecond accuracy triggered by changes in log intensity. In particular, when a change is detected in a pixel, the camera returns an event in the form $e={(x,y,t,p)}$ immediately, where $x,y$ stands for the spatial location, $t$ refers to the timestamp in microseconds, and $p$ is the polarity of the change, either positive or negative. The advantages of high temporal resolution and high dynamic range make event cameras highly effective for analyzing dynamic scenes.

In this work, we systematically review the current state-of-the-art in event-based flow estimation and highlight two critical areas that require further investigation.
First, to the best of our knowledge, {\em there is a notable absence of datasets and methods specifically designed for meshflow in the context of event-based vision}. Existing research primarily focuses on improving the accuracy and robustness of pixel-level event optical flow models. While these efforts have contributed to advancing optical flow estimation, they are still heavily constrained by computational complexity and time-consuming processing, especially when applied to high-resolution images or long video sequences. Additionally, dense optical flow fields often suffer from significant redundancy. In contrast, Meshflow~\cite{liu2016meshflow} offers a promising alternative by segmenting the image plane into a regular grid, with the motion of grid vertices representing the displacement within each cell. This approach significantly reduces computational cost and enhances efficiency, making it highly suitable for real-world applications where performance and scalability are critical.
Second, {\em the density of event data remains a largely underexplored challenge in the field}. The triggering threshold $C$ of event cameras, which is tightly linked to their photosensitive components, is generally fixed during manufacturing. Consequently, different models of event cameras exhibit varying triggering thresholds, leading to significant differences in the event data captured across devices. We conduct a thorough statistical analysis of event data density across four commonly used event camera models, revealing significant differences in both density and resolution, as illustrated in Fig.~\ref{fig:teaser}. These findings underscore the need for more comprehensive methods that account for the varying characteristics of event data and their generalization.

\begin{figure}[t]
    \centering
    \includegraphics[width=\linewidth]{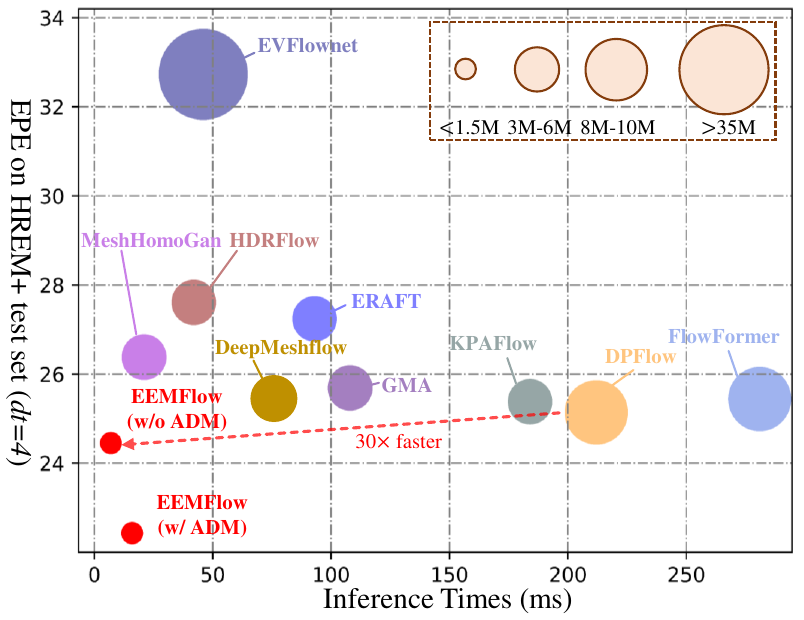}
    \caption{\revised{Comparison of computational overhead and accuracy metrics. The x-axis represents inference time, while the y-axis corresponds to the end-point error. The size of each circle indicates the number of model parameters. Lower values for all metrics are considered better.}}
    \label{fig:computeCost}
\end{figure}

To address these issues, we first create a large-scale \textbf{H}igh-\textbf{R}esolution \textbf{E}vent \textbf{M}eshflow (\textbf{HREM}) dataset, which contains 20k train and 8k test samples. We build 100 virtual scenes in Blender to render the dataset, which can provide accurate physically-based events along with dense meshflow label pairs. Based on the dataset, we further propose an \textbf{E}fficient \textbf{E}vent-based \textbf{M}esh\textbf{F}low (\textbf{EEMFlow}) network to estimate high-resolution meshflow from event data. Unlike recent flow networks relying on recurrent refinement structure~\cite{gehrig2021raft, teed2020raft, jiang2021gma}, our network is developed on an encoder-decoder architecture with multi-scale global optimizing scheme, which can produce full-resolution meshflow with minimal computational overhead. Specifically, our EEMFlow achieves efficiency by employing the lightweight encoder, building cost volume with dilated feature correlation, and using group shuffle convolutions during decoding. \revised{We select recent top-ranked flow networks, including ERAFT~\cite{gehrig2021raft}, KPAFlow~\cite{luo2022kpa}, GMA~\cite{jiang2021gma}, FlowFormer~\cite{huang2022flowformer}, DPFlow~\cite{morimitsu2025dpflow}, as well as HDRFlow~\cite{Xu2024HDRFlow} for HDR imaging, MeshHomoGan~\cite{liu2025meshhomogan} and DeepMeshflow~\cite{liu2025deepmeshflow} for video stabilization, which are trained and evaluated on our dataset. The results demonstrate that our approach achieves state-of-the-art (SOTA) performance while maintaining a fast inference speed of 142.9 FPS, outperforming previous methods by a significant margin (see Fig.~\ref{fig:computeCost}).}

Additionally, we empirically demonstrate that the proposed new pipeline has the capability to effectively handle various motion patterns. Its lightweight design and runtime efficiency further contribute significantly to the field of optical flow estimation.
Specifically, we refine the optical flow progressively during decoding using the coarse to fine residual approach. A \textbf{C}onfidence-induced \textbf{D}etail \textbf{C}ompletion (\textbf{CDC}) module is proposed to preserve motion boundary details during flow upsampling.
We also perform comparative experiments with recent event flow networks~\cite{luo2023learning,liu2023tma,li2023blinkflow} to illustrate its superiority. The enhanced flow network is referred to as {\bf EEMFlow+}, demonstrating impressive performance on the reputable DSEC~\cite{gehrig2021raft} online test benchmark, with the fastest inference speed reaching 39.2 FPS. 

To further investigate the impact of event density on network performance, we simulate event data with varying densities by adjusting the threshold $C$, expanding the HREM dataset into a multi-density dataset, namely \textbf{HREM+}, with densities ranging uniformly from 0.05 to 0.95. Our experiments reveal that networks achieve optimal performance when the input event data has an appropriate density. Based on this finding, we propose a plug-and-play \textbf{A}daptive \textbf{D}ensity Module (\textbf{ADM}), which adjusts the density of input event data to a more suitable range, thereby enhancing the generalization capability and performance of network algorithms across different event datasets. After integrating ADM and fine-tuning, EEMFlow for meshflow estimation achieves an 8\% improvement (Average EPE metrics) on the HREM+ test set, with an inference speed of 62.5 FPS (see Fig.~\ref{fig:computeCost}). \revised{Similarly, with the assistance of ADM, TMA~\cite{liu2023tma}, BFlow~\cite{Gehrig2024BFlow} and EEMFlow+ for optical flow estimation achieve performance improvements of 8\%, 14\% and 10\%, respectively, on the DSEC benchmark (1PE metrics).}

Our contributions are summarized as:
\begin{itemize}
    \item 
    We build \textbf{HREM+}, the first multi-density event meshflow and optical flow dataset, superior in the highest resolution at $1280\times720$, dynamic scenes, complex motion patterns, as well as physically correct accurate events paired with meshflow and optical flow labels. 
    
    \item 
    We propose \textbf{EEMFlow}, which achieves SOTA performances when compared to top-ranked optical flow networks trained on our dataset. \revised{Moreover, it achieves inference speed of 142.9 FPS, which is 30 times faster than the recent state-of-the-art flow method.}

    \item 
    We propose \textbf{CDC}, a confidence-induced detail completion module that empowers EEMFlow to make a notable contribution to the optical flow community. The upgraded model achieves SOTA performance when compared to representative methods, while also boasting the fastest inference speed to date. 

    \item
    We propose a plug-and-play \textbf{ADM} to adjust event density. \revised{In extensive experiments, ADM improves EEMFlow performance by 8\% for meshflow estimation and enhances the performance of TMA, BFlow and EEMFlow+ by 8\%, 14\% and 10\% for optical flow estimation, respectively.}
    
\end{itemize}

\begin{table*}
  \caption{Comparison of Available Datasets: ``Multi Density'' indicates whether the event data in the dataset exhibit varying densities. ``Dense Optical Flow'' and ``Meshflow'' denote the availability of dense optical flow and meshflow labels, respectively.}
  \centering
    \begin{tabular}
    {   >{\arraybackslash}p{2.1cm}| 
        >{\centering\arraybackslash}p{1.8cm}|
        >{\centering\arraybackslash}p{1.8cm}|
        >{\centering\arraybackslash}p{1.8cm}|
        >{\centering\arraybackslash}p{1.8cm}|
        >{\centering\arraybackslash}p{1.8cm}|
        >{\centering\arraybackslash}p{1.8cm}|
        >{\centering\arraybackslash}p{1.8cm}
    }
    \hline
    \multirow{2}{*}{Dataset} &  \multirow{2}{*}{Resolution} & Dynamic  & Extreme  & Multi & Dense Optical  & \multirow{2}{*}{Meshflow} & Motion   \\
    & & Objects & Conditions & Density &  Flow &  & Pattern \\
    \hline
    DVSFLOW~\cite{lin2022dvsvoltmeter} & 240$\times$180 & {\ding{56}} & {\ding{56}} & {\ding{56}} & {\ding{56}} & {\ding{56}} & Rotation \\
    MVSEC~\cite{zhu2018multivehicle} & 346$\times$260 & {\ding{56}} & {\ding{52}} & {\ding{56}} & {\ding{56}} & {\ding{56}} & Drone \\
    DSEC~\cite{gehrig2021dsec}  & 640$\times$480 & {\ding{56}} & {\ding{52}} & {\ding{56}} &{\ding{56}} & {\ding{56}}  & Car \\
    MDR~\cite{luo2023learning}   & 346$\times$260 & {\ding{56}} & {\ding{56}} & {\ding{52}} &{\ding{52}} & {\ding{56}} & Car \\
    BlinkFlow~\cite{li2023blinkflow} & 640$\times$480 & {\ding{52}} &  {\ding{56}} & {\ding{56}} &{\ding{52}} & {\ding{56}} & Random \\
    Ekubric~\cite{wan2023rpeflow} & 1280$\times$720 & {\ding{52}} & {\ding{56}} & {\ding{56}} & {\ding{52}} & {\ding{56}} & Falling \\
    \bf HREM (Ours) & 1280$\times$720 & {\ding{52}} & {\ding{52}} & {\ding{56}} & {\ding{52}} & {\ding{52}} & Random \\
    \bf HREM+ (Ours) & 1280$\times$720 & {\ding{52}} & {\ding{52}} & {\ding{52}} & {\ding{52}} & {\ding{52}} & Random \\
    \hline
    \end{tabular}%
  \label{tab:dataset}%
\end{table*}

In comparison to our earlier conference versions~\cite{luo2023learning, Luo_2024_EEMFlow}, we make the following new contributions. 
First, we statistically analyze the density ranges of datasets captured by different event camera models. These findings underscore the need for more comprehensive methods that account for the varying characteristics of event data and their generalization. 
Second, we provide a detailed description of the HREM dataset synthesis pipeline, including the principles and methods for generating event data. Besides, We extend the HREM dataset to HREM+, which now contains event data with varying densities, providing a more robust basis for studying the impact of event density on network performance. 
Third, we enhance the ADM to better adapt its efficiency with EEMFlow and EEMFlow+. The ADM is lightweight and easily integrable into existing networks, enabling it to adjust input event data density to an optimal range, thus improving the generalization ability and performance of models across diverse event datasets. Two loss functions are designed to train ADM on HREM+.

In addition, we explore the impact of training flow networks with event data of different densities. Experiments are conducted on the test sets of both the MVSEC and DSEC datasets, demonstrating that networks achieve optimal performance when trained with event data of appropriate density. 
Finally, we conduct ablation experiments for ADM. The results show that ADM benefits both meshflow estimation and optical flow estimation. EEMFlow for meshflow estimation achieves an 8\% improvement (Average EPE metrics) on the HREM+ test set. \revised{Similarly, with the assistance of ADM, TMA, BFlow and EEMFlow+ for optical flow estimation achieve performance improvements of 8\%, 14\% and 10\%, respectively, on the DSEC benchmark (1PE metrics). We also verify that ADM can optimize event data across different densities in HREM+, MVSEC and DSEC dataset.}

\section{Related Work} \label{sec:related_work}
\subsection{Image-based Meshflow Warping}
Meshflow is a lightweight and spatially smooth sparse motion field with motions only located at mesh vertices~\cite{liu2016meshflow}. Meshflow is different from dense optical flow, where optical flow estimates motions of every pixel of an image while meshflow only concentrates on the global motion, rejecting motions of any dynamic contents. Meshflow is also different from a global homography, where local motions from nonplanar depth variations can be well reflected. Mesh-based methods proofs to be effective in various applications, such as high dynamic range (HDR) imaging~\cite{yan2017hdr}, burst image denoising~\cite{yang2013denosing}, video denoising~\cite{ren2017meshflow}, image/video stitching~\cite{Nie_2023_ICCV} and video stabilization~\cite{liu2013stable}. It is worth noting that directly downsampling the optical flow may yield flow fields of the same resolution as meshflow, but they differ significantly from meshflow and perform poorly in terms of warping effects~\cite{Luo_2024_EEMFlow}. Direct downsampling ignores motion outliers from different dynamic objects and the global consistency, which can be effectively addressed by meshflow through motion propagation.  

\begin{figure*}[t]
    \centering
    \includegraphics[width=\linewidth]{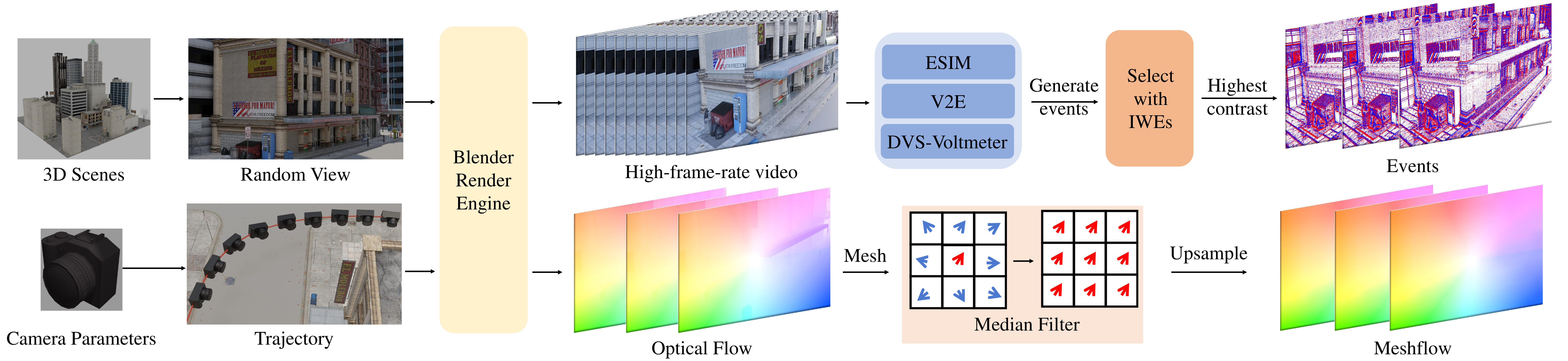}
    \caption{Our data generation pipeline. We generate high-frame-rate video and dense optical flow from a given 3D scene and camera parameters. We then employ three event data simulators to generate events, selecting it as source data which has the highest contrast of IWEs. Finally, we rasterize and median filter the dense optical flow for meshflow as the ground truth.}
    \label{fig:dataset_pipe}
\end{figure*}

\subsection{Event-based Optical Flow Dataset}
Early works synthesize events by thresholding rgb images~\cite{kaiser2016towards} and applying interpolation for high framerate~\cite{gehrig2020video}. However, the timestamp of synthesized events is inaccurate, let alone interpolation artifacts. The DAVIS event camera~\cite{brandli2014240} can capture both images and real events, resulting in some event datasets: DVSFLOW~\cite{lin2022dvsvoltmeter}, MVSEC~\cite{zhu2018multivehicle} and DSEC~\cite{gehrig2021dsec}, based on which EV-Flownet~\cite{zhu2018ev} and ERAFT~\cite{gehrig2021raft} compute sparse optical flow. In this way, however, flows can only locate on sparse event regions~\cite{wan2022DCEI}. Recently, Luo~\emph{et al.}~\cite{luo2023learning} proposed to render the MDR dataset from graphics, but it only contains static scenes. Wan~\emph{et al.}~\cite{wan2023rpeflow} synthesized the Ekubric dataset based on the Kubirc toolbox~\cite{greff2022kubric}, which only includes a single falling motion pattern. Li~\emph{et al.}~\cite{li2023blinkflow} considered non-rigid motions simulating dancing but with lower resolution and without extreme condition scenarios. None of the aforementioned datasets involve the estimation of meshflow, As shown in Tab.~\ref{tab:dataset}.
In this work, we render a comprehensive dataset that can support both meshflow and optical flow estimation, with a higher resolution, rich dynamic scenes, as well as extreme conditions like relatively low light and motion blur.

\subsection{Event-based Optical Flow Estimation}
Optical flow estimation from event cameras has received significant attention in recent years. Early approaches, such as~\cite{benosman2012asynchronous}, could only estimate optical flows at the regions where events are triggered. Recently, deep methods can estimate optical flows from event data, even for the regions without triggered events. For example, EV-FlowNet~\cite{zhu2018ev} learns event and flow labels in a self-supervised manner by minimizing photometric distances of grey images acquired by DAVIS~\cite{brandli2014240}. Various event representations, including EST~\cite{gehrig2019EST} and Matrix-LSTM~\cite{cannici2020Matrix-LSTM}, have been explored, and different network structures, such as SpikeFlowNet~\cite{lee2020spikeflownet}, LIF-EV-FlowNet~\cite{hagenaars2021LIF-EV-FlowNet}, STE-FlowNet~\cite{ding2022STE-FLOWNET}, Li \emph{et al.}~\cite{li2021lightweight}, RTEF~\cite{brebion2021real}, MutilCM~\cite{sun2010secrets}, ERAFT~\cite{gehrig2021raft}, OF-EV-SNN~\cite{cuadrado2023optical}, E2FAI~\cite{Guo25iccv}, Yang \emph{et al.}~\cite{yang2023event}, EVA-Flow~\cite{ye2023towards}, ADMFlow~\cite{luo2023learning}, TamingCM~\cite{paredes2023taming}, MotionPriorCM~\cite{Hamann24motion}, SDformerFlow~\cite{tian2024sdformerflow}, E-FlowFormer~\cite{li2023blinkflow}, IDNet~\cite{wu2024lightweight}, and TMA~\cite{liu2023tma} have been proposed to improve performances. Some methods even use both events and images as input for flow estimation, such as Fusion-FlowNet~\cite{lee2022Fusion-FlowNet}, Pan \emph{et al.}~\cite{pan2020single}, DCEIFlow~\cite{wan2022DCEI} and RPEFlow~\cite{wan2023rpeflow}. In this work, we study a new problem of event-based meshflow estimation, proposing an efficient event-based meshflow network.

\section{Methodology}
\subsection{High-Resolution Event Meshflow Dataset}
We create the high-resolution event meshflow dataset (HREM), consisting of 100 virtual scenes that accurately mimic real-world environments, both indoors and outdoors. In these scenes, we put dynamic objects to simulate intricate object motions. Camera is programmed to track these movements, ensuring a realistic portrayal of motion. The Blender rendering engine was utilized to create high-frame-rate videos and dense optical flow labels. We process the dense optical flow using motion propagation and median filters, enabling the generation of meshflow labels. For event data generation, we employed three advanced simulators: ESIM~\cite{rebecq2018esim}, V2E~\cite{hu2021v2e}, and DVS-Voltmeter~\cite{lin2022dvsvoltmeter}. The simulator that provided the highest contrast in warped events images was chosen for our dataset. The overview of our data generation pipeline is shown in Fig.~\ref{fig:dataset_pipe}. As an extension, we adjust the triggering threshold 
$C$ of the event data simulators to generate event sequences with varying densities from the same scene and time interval. This approach emulates real-world scenarios where different models of event cameras capture data at varying densities.

\subsubsection{Rendering Platform}  
Our rendering platform is built on Blender, an open-source 3D development tool, enabling the import of custom 3D scenes and simulation of complex camera movements to generate high-quality, large-scale data.  

\textbf{3D Scenes.} To align the distribution of generated event data with real-world captures, we curate a diverse collection of realistic virtual scenes from Blender. Dynamic entities with realistic motions (e.g., walking, driving, cycling) are integrated into these scenes using Blender plugins. This process result in a dataset of 100 virtual scenes, comprising 40 indoor and 60 outdoor environments.  

\textbf{Camera Trajectory.} Camera movements are simulated using six degrees of freedom (6-DoF), with 3-DoF for translation and 3-DoF for rotation. For scenes with dynamic objects, the camera is programmed to follow the entities along a predefined trajectory, ensuring coherent motion capture. To avoid collisions and maintain visibility within the scene, we utilize PyBullet~\cite{coumans2016pybullet}, a physics engine, to generate camera paths. Starting with specified initial and final positions, random translational and rotational perturbations are added to create a smooth trajectory function \(\Gamma(t)\), which outputs the camera's position and orientation at time \(t\), denoted as \(P(t) = [x(t), y(t), z(t), r(t)]^T\).  


\subsubsection{Generation of Optical Flow and Meshflow}\label{sec:meshflow_form_flow}
Given two timestamps \(t_i\) and \(t_j\), the positions and orientations \(P(t_i)\) and \(P(t_j)\) of the camera are derived from \(\Gamma(t)\). Using the specialized render pass of Blender for auxiliary ground truth, forward and backward optical flow \((F_{i \to j}, F_{j \to i})\) are computed. Meshflow \(MF\) is then extracted from the optical flow \(F\).  

\textbf{Motion Propagation.} 
Given a dense optical flow $F$, we place a uniform mesh of 16$\times$16 regular cells on its image plane and then select the motion of the middle point $p$ in each cell as the local motion $v_p$. Since the vertices of the mesh near point $p$ should have a similar motion to $v_p$, we define a rectangle that covers 3$\times$3 cells centered at $p$, and assign to all the vertices within the rectangle, ensuring the propagation of the local motion $v_p$ across the image plane. 

\begin{figure}[t]
    \centering
    \includegraphics[width=\linewidth]{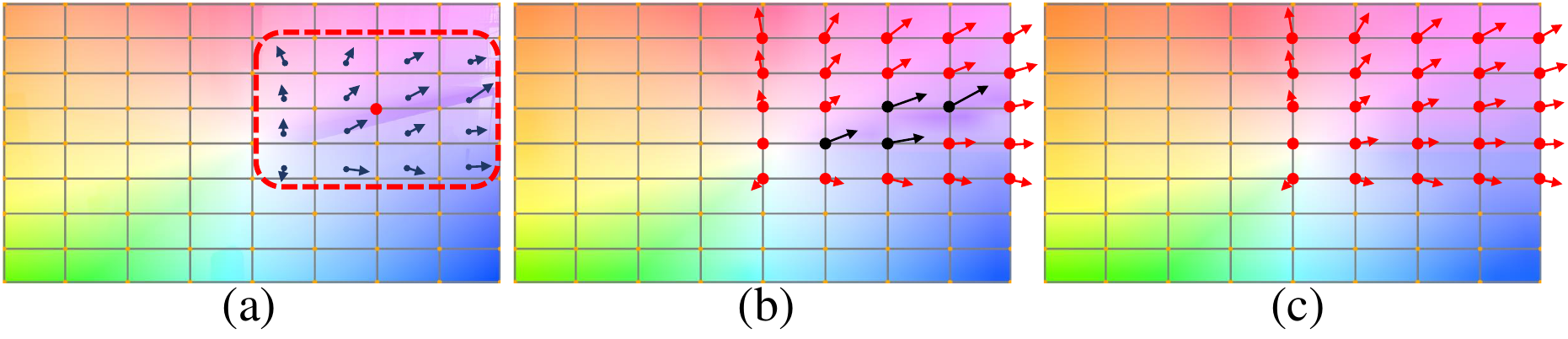}
    \vspace{-2em}
    \caption{The process of generating meshflow from dense optical flow. (a) Propagate the motion vector of each grid center to the grid vertices. (b) Apply median filter $f_1$ to multiple motion vectors of each vertex to select the most appropriate motion for that vertex. (c) Use median filtering $f_2$ to smooth the motion field in the mesh grid. For ease of visualization, we present the $8 \times 8$ grid mesh in this paper.}
    \label{fig:meshflow}
    \vspace{-.5em}
\end{figure}

\textbf{Median Filters.} 
The local motions of the middle points in all cells are propagated to their nearby mesh vertices, resulting in each vertex potentially receiving multiple motion vectors. To select the most appropriate motion vector for a given vertex, we apply a median filter $f_1$ to filter the candidate motions. The response of the filter is then assigned to the corresponding vertex. The median filter is a widely used technique in optical flow estimation and has been shown to produce high-quality flow estimates~\cite{sun2010secrets}. Therefore, we use the median filter for sparse motion regularization. However, due to dynamic objects, the motion field may contain noise and needs to be spatially smoothed. To address this issue, we apply another median filter $f_2$ that covers 3$\times$3 cells neighborhood to suppress the noise in the motion field. This second median filter produces a spatially-smooth sparse motion field, which is called as meshflow. Ultimately, we use the meshflow (generally 16$\times$16) as the label, and upsample it to the full image resolution for intuitive display and alignment applications. \revised{For meshflow representing global motion, bilinear interpolation is sufficient for upsampling, and experiments confirm that bilinear interpolation is optimal.} The whole pipeline is as illustrated in Fig.~\ref{fig:meshflow}.

\begin{figure*}[t]
    \centering
    \includegraphics[width=1\linewidth]{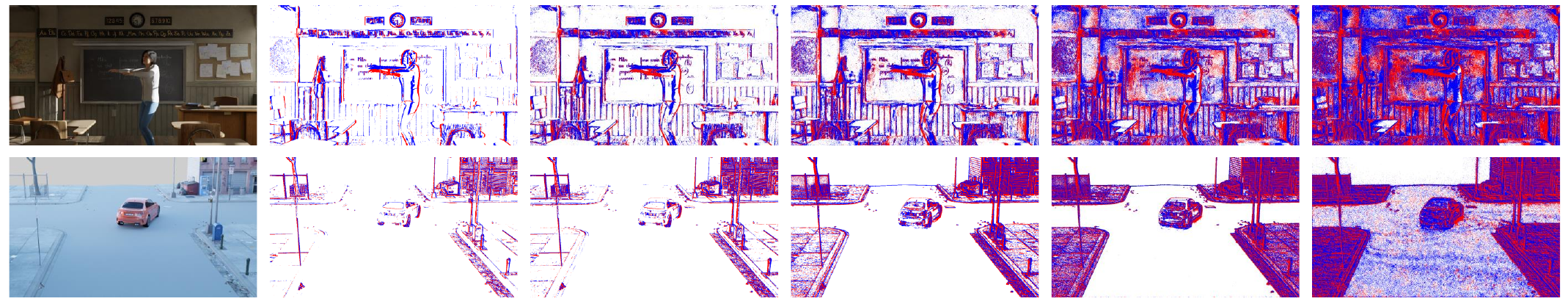}
    \vspace{-1.5em}
    \caption{Illustration of two dynamic scenes from HREM+ dataset. The first column shows the reference image, while columns two to five visualize event data at different densities, with density increasing from left to right.}
    \label{fig:dataset_multidensity}
\end{figure*}

\begin{figure*}[t]
    \centering
    \includegraphics[width=\linewidth]{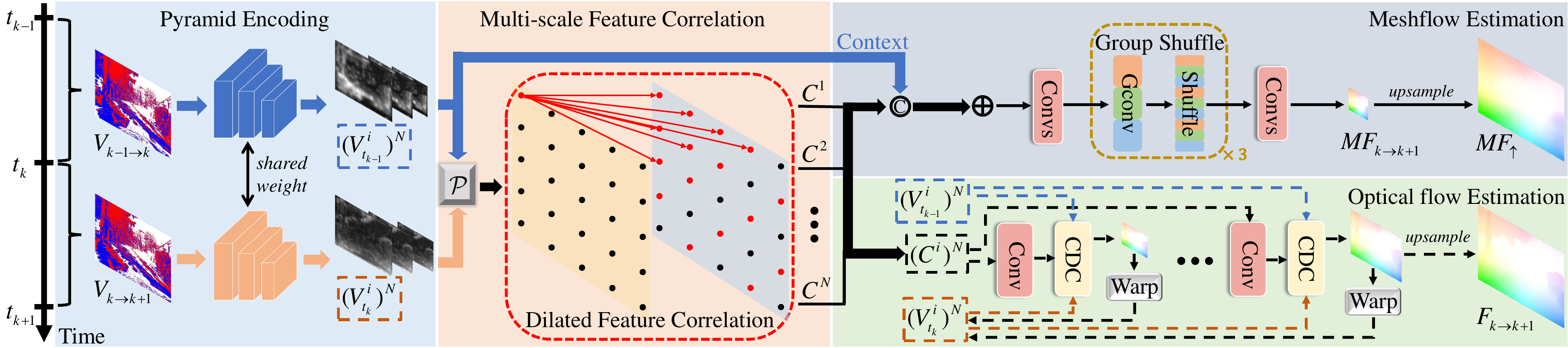}
    \vspace{-1.5em}
    \caption{Our proposed network architecture. We employ pyramid encoders to extract multi-scale features from $V_{k-1 \to k}$ and $V_{k \to k+1}$, then use dilated feature correlation to compute the cost volume between each layer of features, followed by decoding to output the results. For meshflow estimation, we utilize the decoders with group shuffle convolutions to output predictions $MF_{k \to k+1}$, upsampled to $MF\uparrow_{k \to k+1}$. For optical flow estimation, we refine flows using a coarse-to-fine residual approach and confidence-induced detail completion module, finally outputing the optical flow $F_{k \to k+1}$. } 
    \label{fig:networks}
\end{figure*}

\subsubsection{Generation of Multi-Density Event Data}  
We use Blender to generate high-frame-rate videos, where the frame rate correlates with the optical flow displacement, and synthesize event data by capturing continuous brightness variations from the videos.  

\textbf{High-Frame-Rate Video.}  
The camera trajectory \(\Gamma(t)\) is sampled between timestamps \(t_i\) and \(t_j\) to produce camera positions and poses \(\{P(\tau_k)^N\}, k \in [0, N]\), which are then used in Blender's rendering pass to generate high-frame-rate video frames \(\{I(\bm{u},\tau_k)^N\}, k \in [0, N]\). Following ESIM~\cite{rebecq2018esim}, we adopt an adaptive sampling strategy to ensure the maximum pixel displacement between successive frames \(I(\bm{u},\tau_k)\) and \(I(\bm{u},\tau_{k+1})\) is under one pixel. The sampling interval \(\Delta \tau_k\) is defined as:  
\begin{equation}
\begin{aligned}
\Delta \tau_{k} &= \tau_{k+1} - \tau_{k} \\
&= (\max_{\bm{u}} \max \{ \left \| F_{{{k-1}} \to {{k}}} \right \| , \left \| F_{{k} \to {k-1}} \right \|\})^{-1},
\end{aligned}
\label{sampling}
\end{equation}
where \(F_{k-1 \to k}\) and \(F_{k \to k-1}\) are the optical flows between consecutive frames.  

\textbf{Event Simulator.}  
Event cameras asynchronously respond to changes in the logarithmic intensity signal \(L(\bm{u_e}, t_e) \doteq \ln(I(\bm{u_e}, t_e))\) at each pixel. An event is triggered at pixel \(\bm{u_e} = (x_e, y_e)\) and time \(t_e\) when the intensity change since the last event exceeds a threshold \(\pm C (C > 0)\):  
\begin{equation}
     L(\bm{u_e},t_e) - L(\bm{u_e},t_e-\Delta t) \ge  p_e C,
\label{event_generation}
\end{equation}
where \(t_e\) is the event timestamp, \(\Delta t\) is the time since the last event at \(\bm{u_e}\), and \(p_e \in \{-1, +1\}\) is the event polarity. The event sequence \(E(t_i, t_j)\) is defined as \(\{(\bm{u_e}, t_e, p_e)^{N_e}\}, e \in [1, N_e]\), representing \(N_e\) events between \(t_i\) and \(t_j\).  

Based on this ideal generation model, we synthesize event data using brightness variation signals from high-frame-rate videos. Three state-of-the-art simulators are currently available: ESIM~\cite{rebecq2018esim}, V2E~\cite{hu2021v2e}, and DVS-Voltmeter~\cite{lin2022dvsvoltmeter}. A summary of their differences is provided in Tab.~\ref{tab:simulator}.  
ESIM can be considered an early simulator that is simple and straightforward, providing the fastest runtime. V2E takes into account lighting conditions and is suitable for scenes with low illumination. DVS-Voltmeter considers the impact of sensor operating temperature and works more similar to the real DAVIS event camera when simulating data over long durations.

\begin{table}[t]
  \caption{The difference of three state-of-the-art event simulators. 10k ev./s means 10k events are generated per second.}
  \centering
    \begin{tabular}
    {   >{\arraybackslash}p{2.5cm}| 
        >{\centering\arraybackslash}p{1.5cm}|
        >{\centering\arraybackslash}p{1.5cm}|
        >{\centering\arraybackslash}p{1.5cm}
    }
    \hline
    \multirow{2}{*}{Simulators} &Low &Temperature &Speed \\
    &Illumination &Noise &(10k ev./s) \\
    \hline
    ESIM~\cite{rebecq2018esim} &\ding{56} & \ding{56} & 170 \\
    V2E~\cite{hu2021v2e}  &\ding{52}  &\ding{56}  & 26 \\
    DVS-Voltmeter~\cite{lin2022dvsvoltmeter} &\ding{56} &\ding{52}  & 29 \\
    \hline
    \end{tabular}
  \label{tab:simulator}
\end{table}

\textbf{Event Data Generation.} 
We generate three event sequences for the same high-frame-rate video using different simulators and select the sequence with the highest contrast in the Image of Warped Events (IWE)~\cite{Gallego2018Contrast,Back_to_Basics, paredes2023taming, shiba2024secrets} for inclusion in the dataset.  

Given an event sequence \(E(t_{i}, t_{j})\) and the corresponding optical flow \(F_{i \to j}\), each event \((\bm{u_e}, t_e, p_e)\) is warped to a reference time \(t_{ref}\), producing
\begin{equation}
\bm{u^{'}_e} = \bm{u_e} + \frac{t_{ref} - t_e}{t_j - t_i} F_{i \to j}(\bm{u_e}),
\label{eventwarp}
\end{equation}
These warped events are aggregated to form an IWE \(I(\bm{u}, t_{ref})\):  
\begin{equation}
I(\bm{u}, t_{ref}) = \sum_{e}^{N_e} \delta(\bm{u} - \bm{u_e^{'}}),
\label{IWE}
\end{equation}
and the contrast of IWE is computed as:  
\begin{equation}
\mathrm{Var}(I(\bm{u}, t_{ref})) = \frac{1}{|\Omega|} \int_{\Omega} |I(\bm{u}, t_{ref}) - \mu_I|^2 d\bm{u},
\label{contrast}
\end{equation}
where \(\mu_I\) is the mean IWE value. To evaluate contrast, we warp events to \(t_i\) and \(t_j\), compute their respective contrasts, and sum them:  
\begin{equation}
\mathrm{Var} = \mathrm{Var}(I(\bm{u}, t_{i})) + \mathrm{Var}(I(\bm{u}, t_{j})).
\label{Var}
\end{equation}
The event sequence with the highest contrast \(\mathrm{Var}\) is selected for the dataset.  

To simulate event data with varying densities, we adjust the event generation threshold \(C\). Since event streams are often converted into representations like voxel grids~\cite{zhu2019unsupervised, stoffregen2020reducing, gehrig2021raft} before being processed by deep networks, we measure event density as the percentage of valid pixels (pixels with at least one event) in the voxel representation:  
\begin{equation}
V(\bm{u_e}, b) = \sum_{e=0}^{N}p_{e} \max(0, 1-|b-\frac{t_e-t_0}{t_N-t_0}(B-1)|),
\label{reprentation}
\end{equation}
\begin{equation}
D = \frac{1}{HW}\sum_{e=1}^{N} \varepsilon (\sum_{b=0}^{B-1}|V(\bm{u_e}, b)|),
\varepsilon(x)=\left\{\begin{matrix}1 , x>0\\0,x \le 0\end{matrix}\right.,
\label{density}
\end{equation}
Here, \(V(\bm{u_e}) \in \mathbb{R}^{B \times H \times W}\) is the voxel grid representation of the event stream, \(B\) is the number of temporal bins (typically 5 or 15), and \(D\) represents event density.  

In practice, different event cameras and scenarios produce data with varying densities. To address this, we adjust the event generation threshold 
\(C\) to extend HREM into HREM+, which contains event data uniformly distributed across the density range [0.05, 0.95]. This ensures compatibility with datasets of different density ranges. Visualization examples of multi-density event data from the extended HREM+ are shown in Fig.~\ref{fig:dataset_multidensity}.

\subsection{Estimation for Meshflow and Optical Flow}
Based on~\cite{gehrig2021raft},  we estimate the meshflow $MF_{k \to k+1}$ and optical flow $F_{k \to k+1}$ from two consecutive event sequences $E(t_{k-1},t_{k})$ and $E(t_{k},t_{k+1})$. The overall architecture of our network is in Fig.~\ref{fig:networks}. \revised{Following}~\cite{zhu2019unsupervised}, we convert the input event sequences $E(t_{k-1},t_{k})$ and $E(t_{k},t_{k+1})$ to the 3D volumns $V_{k-1 \to k}$ and $V_{k \to k+1}$.


\subsubsection{Efficient Event-based Meshflow Network}
We propose EEMFlow to directly output results of the same resolution as the ground truth meshflow $MF_{GT}$ for supervised regression, fully leveraging the advantage about low-parameter and high-motion-information of meshflow.

\textbf{Overall Structure of Meshflow Estimation.} Since meshflow focuses more on global large motion rather than local detailed motion, the EEMFlow we designed does not require excessively deep network layers or refinement operations from coarse to fine. Firstly, EEMFlow employs a three-level pyramid feature encoder to extracting features $(V^i_{t_{k-1}})^N$ and $(V^i_{t_{k}})^N$ from $V_{k-1 \to k}$ and $V_{k \to k+1}$, the convolutional layers within the $i$-th level share weights for $V^i_{t_{k-1}}$ and $V^i_{t_{k}}$. Secondly, EEMFlow utilizes multi-scale feature correlation to builds the cost volumes for meshflow estimation. Features $(V^i_{t_{k-1}})^N$ and $(V^i_{t_{k}})^N$ undergo average pooling operation $\mathcal{P}$ to the same resolution ($1/64$ resolution of $V_{k \to k+1}$), and then use correlation to capture relative motion information and output cost volumes $(C^i)^N$. Specifically, we employ the dilated feature correlation (DFC) to increase the search area while reducing computational parameters. Finally, we stack the cost volumes $(C^i)^N$ and features $(V^i_{t_{k-1}})^N$ after pooling, fuse them with a weighted-sum operation, and then feed them into the decoders to regress the meshflow $MF_{k \to k+1}$ at the same resolution as $MF_{GT}$. Inspiring by~\cite{zhang2018shufflenet}, we replace the conventional convolutions with group shuffle convolutions, which leads to efficient computation while maintaining high accuracy. 

\begin{figure}[t]
\centering
\includegraphics[width=\linewidth]{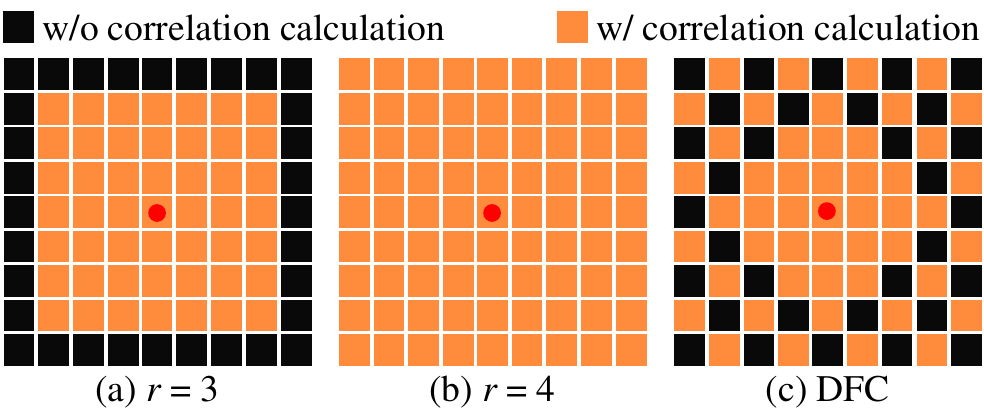}
\caption{\revised{Different search grid for computing correlation. The yellow cell represents the neighboring pixels that need to be involved in the correlation calculation, while the black cells represent what does not participate in the calculation.}}
\label{fig:correlation_searchgrid}
\end{figure}

\textbf{Dilated Feature Correlation.}
We use the inner product to calculate correlation between $V^i_{t_{k-1}}$ and $V^i_{t_{k}}$ for meshflow estimation :
\begin{equation}
C^i(\bm{u}, \bm{d}) = V^i_{t_k} (\bm{u}) \cdot V^i_{t_{k-1}}(\bm{u}+\bm{d}) / M, \bm{d} \in \mathcal{N},
\label{correlation}
\end{equation} 
where $\bm{u}$ represents the spatial coordinates on $V^i_{t_{k}}$, $\mathcal{N}$ represents the search grid of coordinate $\bm{u}$ in feature $V^i_{t_{k-1}}$, $M$ represents the number of elements in $\mathcal{N}$, $\bm{d}$ represents the offset coordinates of the elements in $\mathcal{N}$, and $\cdot$ represents the inner product calculation. Many methods like~\cite{sun2018pwc, Hur:2019:IRR} simply define $\mathcal{N}$ as a square range of size $(2r+1) \times (2r+1)$ and observe that increasing the radius $r$ can reduce errors but increase computational overhead. We propose Dilated Feature Correlation (DFC) that samples densely around the center and sparsely at farther distances, thereby reducing computation while enabling larger radius cost volumes, as Eq.~\ref{search_grid}:
\begin{equation}
\mathcal{N}(d_x,d_y)=\begin{cases}0,\mathrm{if}~| d_x | +| d_y |=2k,k\in[2,r], \\1,\mathrm{others},\end{cases}
\label{search_grid}
\end{equation} 
where $\bm{d}=(d_x, d_y)$ represents the relative position coordinates within the neighborhood $\mathcal{N}$, where $\mathcal{N}(d_x,d_y)=1$ indicates the computation of correlation, and $\mathcal{N}(d_x,d_y)=0$ signifies no computation. \revised{As shown in Fig.~\ref{fig:correlation_searchgrid}, our DFC achieves large correlation search grids while reducing computational overhead.}

\begin{figure}[t]
    \centering
    \includegraphics[width=\linewidth]{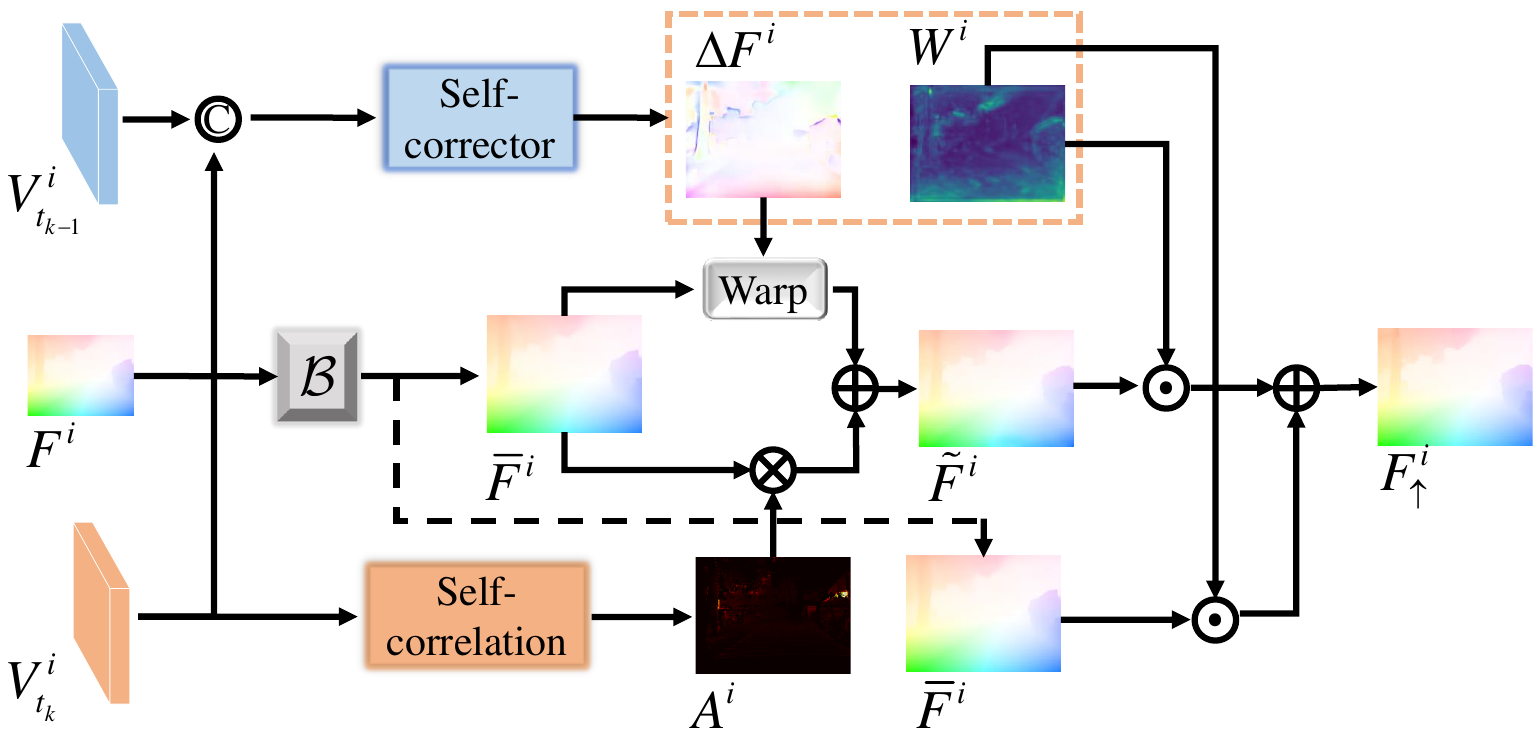}
    \caption{The structure of CDC. CDC employs self-corrector based on a dense convolutional network and self-correlation based on a self-attention mechanism to correct the flow obtained from bilinear upsampling. $\mathcal{B}$ represents bilinear upsampling.}
    \label{fig:CDC_submule}
\end{figure}

\subsubsection{Event-based Optical Flow Network}
Based on EEMFlow, we make some improvements for accurate estimation of optical flow $F_{k \to k+1}$, upgrading it to EEMFlow+. Since optical flow focuses more on local motion and pays attention to object edge details, we employ the coarse to fine residual approach to progressively refine the flow, which can be expressed as Eq.~\ref{coarse_to_fine}.
\begin{equation}
F^{i+1}=\mathrm{Conv}^i(\mathcal{C}(V^i_{{t_{k-1}}},~\mathcal{W}(V^i_{t_k},~F^i_\uparrow )) ) +F^i_\uparrow ,
\label{coarse_to_fine}
\end{equation} 
where $\mathcal{C}(\cdot , \cdot)$, $\mathcal{W}(\cdot , \cdot)$ and $\uparrow$ donates our dilated feature correlation, the warping operation and upsampling, respectively. We employ the pyramid decoders for optical flow, thus $F^i$ is the output flow of the $i$-th decoder, $F^{i+1}$ and $\mathrm{Conv}^i$ are respectively the output flow and convolutions in the $i+1$-th decoder. The most noteworthy aspect is upsampling $F^{i}$ to $F^{i}_\uparrow$. Many methods~\cite{dosovitskiy2015flownet,zhu2018ev,sun2018pwc,Hur:2019:IRR} use bilinear interpolation for upsampling, but this can lead to the mixing of incorrect motions at object edges, resulting in blurring. Therefore, we propose confidence-induced detail completion module for upsampling to enhance edge details.

\textbf{Confidence-induced Detail Completion Module.} 
We propose the confidence-induced detail completion module (CDC) to eliminate the blurring of object edges caused by the mixing of multiple motions at the junctions of different movements during upsampling. The detailed structure of our CDC is shown in Fig.~\ref{fig:CDC_submule}. Given a small-scale flow $F^i$ from the $i$-th level, our CDC first generates an initial flow $\bar{F}^i$ for the $i+1$-th level through bilinear interpolation, and then arranges the self-corrector and self-correlation branches to correct it. Self-corrector is based on a dense convolutional network with a five-layer structure. It captures motion information within the edge neighborhood through dense convolution from the concatenated feature $V^i_{t_{k-1}}$ and $V^i_{t_{k}}$, outputting the corrected flow $\Delta F^i$ and the corrected confidence map $W^i$. Self-correlation is based on a self-attention mechanism, using a large receptive field to find the fine regions in features $V^i_{t_{k}}$ that are identical to the motion of the error region in $\bar{F}^i$. It outputs self-attention weights $A^i$, multiplied with the initial flow $\bar{F}^i$. With the corrected flow $\Delta F^i$  and self-attention weights $A^i$, we can generate the fine flow $\tilde{F}^i$ :
\begin{equation}
\tilde{F}^i= \alpha \mathcal{W}(\bar{F}^i, \Delta F^i) + (1 - \alpha) (A^i \otimes \bar{F}^i),
\label{correct_flow}
\end{equation} 
where $\mathcal{W}(\cdot , \cdot)$ means the warping operation, $\otimes$ donates multiplication and $\alpha \in [0,1]$ is the weight coefficient. \revised{We empirically find that $\alpha = 0.6$ yields the best results.} 

Next, we identify error-prone object edge areas based on the corrected confidence map $W^i$, as weights to fuse the initial flow $\bar{F}^i$ and the fine flow $\tilde{F}^i$,  obtaining the final corrected flow $F^i_\uparrow$:
\begin{equation}
F^i_\uparrow= W^i \odot \bar{F}^i + (1 - W^i) \odot \tilde{F}^i,
\label{select_correct}
\end{equation} 
where $\odot$ donates the element-wise multiplier. \revised{The confidence map $W^i$, learned during training, effectively highlights error-prone regions, particularly along object edges.}

\subsubsection{Loss Function}
For both meshflow and optical flow estimation, we use L1 loss for supervised regression during training. Our EEMFlow, used for estimating meshflow, directly outputs results at the same resolution as the meshflow GT, allowing for direct loss calculation. Similarly, the results outputted by our EEMFlow+ for optical flow estimation are at the same resolution as the network input, enabling direct calculation with the optical flow GT.

\begin{figure}[t]
    \centering
    \includegraphics[width=\linewidth]{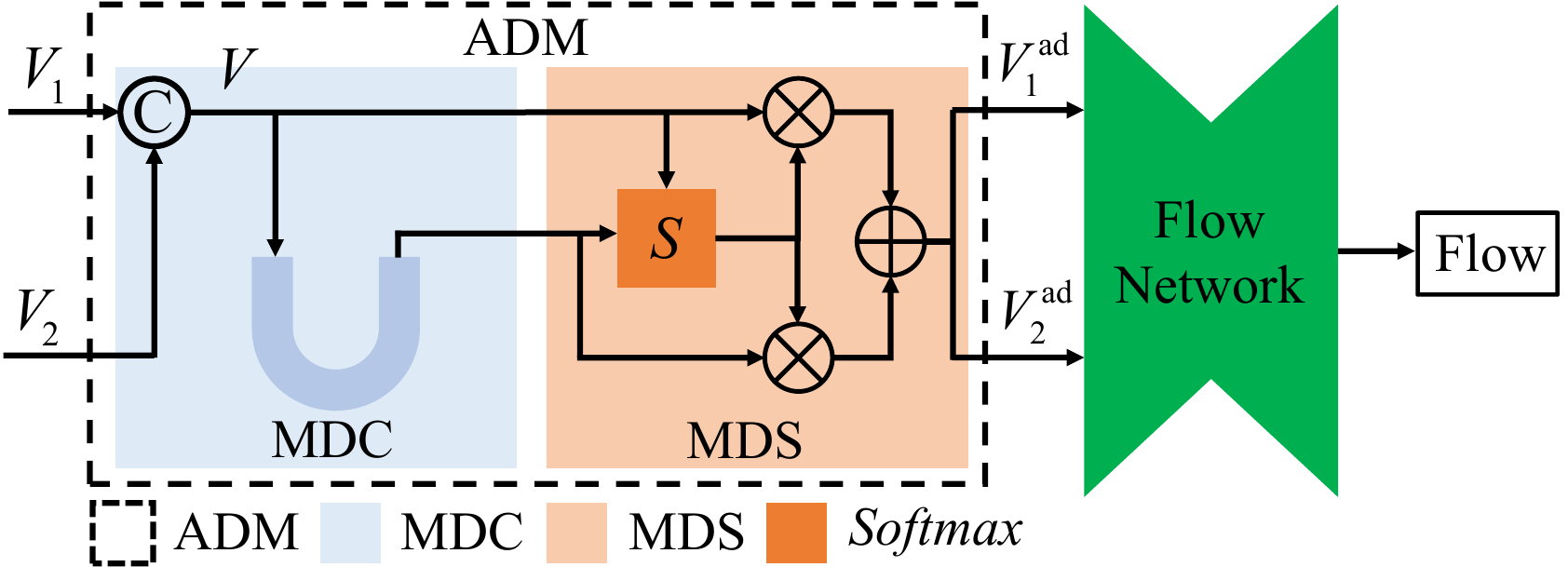}
    \caption{The structure of ADM. We design a plug-and-play Adaptive Density Module (ADM) to transform input event representations $V_1$ and $V_2$ into 
    $V_1^{\mathrm{ad}}$ and $V_2^{\mathrm{ad}}$ with suitable density for optical flow estimation. }
    \label{fig:adm_network}
\end{figure}

\subsection{Extension of Adaptive Density Module}
Event-based optical flow estimation involves predicting dense optical flow \(F_{k-1 \to k}\) from consecutive event sequences \(E(t_{k-1}, t_k)\) and \(E(t_k, t_{k+1})\). The same approach applies to Meshflow. In this paper, we observe that networks perform better on event sequences with appropriate density, compared to those with excessively sparse or dense events. Motivated by this, we propose a plug-and-play Adaptive Density Module (ADM) that adjusts the input event stream to a density optimized for optical flow or meshflow estimation, shown in Fig.~\ref{fig:adm_network}. The ADM transforms input event representations \(V_1\) and \(V_2\) into adjusted representations \(V_1^{\mathrm{ad}}\) and \(V_2^{\mathrm{ad}}\), which are then used by a flow estimation network.

\subsubsection{Adaptive Density Module}
The ADM consists of two sub-modules: the Multi-Density Changer (MDC) and the Multi-Density Selector (MDS). The MDC globally adjusts the density of input event representations across multiple scales, while the MDS selects the optimal pixel-wise density for optical flow estimation.

The MDC adopts an encoder-decoder architecture with three levels. It generates multiscale transformed representations \(V_3^{\mathrm{MDC}}, V_2^{\mathrm{MDC}}, V_1^{\mathrm{MDC}}\) (also denoted as \(V_{\mathrm{out}}^{\mathrm{MDC}}\)) from concatenated input event representations \(V\). Three encoding blocks extract multi-scale features, followed by three decoding blocks and two feature fusion blocks. To keep the module lightweight, each block contains only two \(3 \times 3\) and one \(1 \times 1\) convolutional layers. To achieve lightweight design, we configure each block with only one convolutional layer, followed by a batch normalization layer and an activation function layer.

The MDS module adapts the density transformation by selecting and fusing \(V_{\mathrm{out}}^{\mathrm{MDC}}\) and \(V\). First, the two representations are concatenated, and two convolutional layers generate selection weights via softmax. The weights are then used to fuse \(V_{\mathrm{out}}^{\mathrm{MDC}}\) and \(V\), resulting in the adjusted event representations \(V_1^{\mathrm{ad}}\) and \(V_2^{\mathrm{ad}}\), which are fed into an existing optical flow network.

Using our HREM dataset with multi-density events, we train ADM with an event representation of moderate density as the ground truth (\(V^\mathrm{GT}\)).

\subsubsection{Multi-Density Loss}
For the MDC module, we use a multi-scale loss:  
\begin{equation}
L_{\mathrm{MDC}}=\sum_{k=1}^{3}\sqrt{(V_k^\mathrm{MDC}-V_k^\mathrm{GT})^2+\xi^2},
\label{MDCloss}
\end{equation} 
where \(\xi = 10^{-3}\) is a constant, and \(V_k^{\mathrm{MDC}}\) is the output of the \(k\)-th MDC level, while \(V_k^{\mathrm{GT}}\) is the downsampled ground truth.

For the MDS module, we calculate the density difference between \(V^{\mathrm{ad}}\) and \(V^{\mathrm{GT}}\):  
\begin{equation}
L_{\mathrm{MDS}}=\left \| D(V^\mathrm{ad}) - D(V^\mathrm{GT}) \right \|_1,
\label{MDSloss}
\end{equation} 
where \(D\) computes density as described in Eq.~(\ref{density}).

The total loss function for training the entire pipeline is:  
\begin{equation}
L_{total} = \lambda_1 L_{\mathrm{MDC}} + \lambda_2 L_{\mathrm{MDS}} + L_{\mathrm{Flow}},
\label{totalloss}
\end{equation} 
where \(L_{\mathrm{Flow}}\) represents the flow network loss. We empirically set \(\lambda_1 = 0.1\) and \(\lambda_2 = 10\).

\begin{table*}[t]
  \centering
    \vspace{-0.5cm}
  \caption{\revised{Quantitative comparison of our EEMFlow with other advanced flow networks on our HREM+ dataset. The evaluation metric used is End-Point Error (EPE). ``Parameters'' and ``Time'' respectively indicate the network parameter count and inference time. We evaluate four test sub-sequences (outdoor\_slow, outdoor\_fast, indoor\_slow, indoor\_fast) and report the average EPE across all sequences (“Avg” column, lower is better). All baselines are adapted from mesh warping methods (video stabilization and HDR imaging) or recent optical flow estimation methods by replacing their inputs with event representations. $\Delta P$ and $\Delta T$ represent the change in parameter count and inference time relative to ERAFT~\cite{gehrig2021raft} for other networks. Smaller values are desirable for all metrics. We highlight the best results in \textcolor{red}{red} and the second-best results in \textcolor{blue}{blue}.}}
    \begin{tabular}
    {   >{\arraybackslash}p{4.0cm}| 
        >{\centering\arraybackslash}p{1.5cm} %
        >{\centering\arraybackslash}p{1.5cm}| %
        >{\centering\arraybackslash}p{1.2cm} 
        >{\centering\arraybackslash}p{1.2cm}| 
        >{\centering\arraybackslash}p{1.2cm} 
        >{\centering\arraybackslash}p{1.2cm}| 
        >{\centering\arraybackslash}p{1.2cm} 
    }
    \hline
    Method & Parameters & Time & \multicolumn{2}{c|}{Outdoor} & \multicolumn{2}{c|}{Indoor} &  \multirow{2}{*}{Avg} \\
    $dt=1$ & (M) & (ms) & Slow & Fast & Slow & Fast & \\
    \hline
    EV-FlowNet~\cite{zhu2018ev} & 38.2 & 46 & 3.55  & 16.16  & 2.93  & 11.65   & 8.57   \\
    PWCNet~\cite{sun2018pwc} & {3.36} & {42} & 3.91  & 14.49   & 2.86  & 11.89 & 8.29  \\
    ERAFT~\cite{gehrig2021raft} & 5.27 & 93 & 4.15   & 13.32  & 2.91 & 10.34 & 7.68   \\
    SKFlow~\cite{sun2022skflow} & 6.28 & 145 & 3.76  & 11.78  & 7.24  & 8.81  & 7.24  \\
    GMA~\cite{jiang2021gma} & 5.89 & 108 & 2.18  & 12.07  & 2.02  & 9.34   & 6.40  \\
    KPAFlow~\cite{luo2022kpa} & 6.00 & 184 & \textcolor{blue}{2.03}  & 12.25  & {1.95} & 9.02 & 6.31  \\
    FlowFormer~\cite{huang2022flowformer} & 9.87 & 281 & {2.06}  & {11.71}  & {1.88} & {8.66}  & {6.08} \\
    HDRFlow~\cite{Xu2024HDRFlow} & 3.16 & 32 & 3.82 & 14.12 & 2.86 & 10.58 & 7.85   \\
    MeshHomoGan~\cite{liu2025meshhomogan} & {2.66} & {21} & 3.62 & 12.97 & 2.74 & 9.52 & 7.21   \\
    DeepMeshflow~\cite{liu2025deepmeshflow} & 4.19 & 76 &2.32 & 11.21 & 2.23 & 9.14 & 6.23   \\
    DPFlow~\cite{morimitsu2025dpflow} & 10.0 & 212 & \textcolor{red}{1.96} & 12.02 & \textcolor{red}{1.81} & {8.82} & {6.15}  \\
    EEMFlow (Ours) & \textcolor{red}{1.24} & \textcolor{red}{7} & 2.42  & \textcolor{blue}{9.09} & 2.00  & \textcolor{blue}{8.46}  & \textcolor{blue}{5.50}  \\
    EEMFlow$_{\rm{ADM}}$ (Ours) & \textcolor{blue}{1.48} & \textcolor{blue}{16} & 2.24 & \textcolor{red}{8.15} & \textcolor{blue}{1.87} & \textcolor{red}{7.94} & \textcolor{red}{5.05} \\
    \hline
    Method & $\Delta P$ & $\Delta T$ & \multicolumn{2}{c|}{ Outdoor } & \multicolumn{2}{c|}{Indoor} & \multirow{2}{*}{Avg} \\
    $dt=4$ & (M) & (ms) & Slow & Fast & Slow & Fast & \\
    \hline
    EV-FlowNet~\cite{zhu2018ev} & +624\% & +51\% & 18.25  & 49.32  & 16.16  & 47.19  & 32.73 \\
    PWCNet~\cite{sun2018pwc} & {-36\%} & {-55\%} & 16.40   & 46.17  & 14.49 & 40.90 & 29.49  \\
    ERAFT~\cite{gehrig2021raft} & 0\% & 0\% & 15.21  & 40.83  & 13.32  & 39.61  & 27.24  \\
    SKFlow~\cite{sun2022skflow} & +19\% & +56\% & 14.93  & 39.24  & {11.71} & 39.22 & 26.28  \\
    GMA~\cite{jiang2021gma} & +11\% & +16\% & 14.13  & 38.89  & 12.07  & 37.68   & 25.69  \\
    KPAFlow~\cite{luo2022kpa} & +14\% & +99\% & 14.04  & {38.03}  & 12.25 & {37.20} & {25.38} \\
    FlowFormer~\cite{huang2022flowformer} & +88\% & +202\% & {13.89} & {38.55} & {10.77} & 38.53 & 25.44 \\
    HDRFlow~\cite{Xu2024HDRFlow} & -40\% & -66\% & 14.99 & 41.24 & 13.28 & 40.92 & 27.61  \\
    MeshHomoGan~\cite{liu2025meshhomogan} & {-50\%} & {-77\%} & 14.12 & 40.23 & 13.07 & 38.11 & 26.38  \\
    DeepMeshflow~\cite{liu2025deepmeshflow} & +20\% & +18\% & 13.96 & 38.14 & 12.68 & 36.96 & 25.44 \\
    DPFlow~\cite{morimitsu2025dpflow} & +90\% & +128\% & \textcolor{blue}{13.82} & 39.26 & \textcolor{blue}{10.68} & {36.78} & 25.14  \\
    EEMFlow (Ours) & \textcolor{red}{-76\%} & \textcolor{red}{-92\%} & {13.97}  & \textcolor{blue}{37.33} & 12.09 & \textcolor{blue}{34.39} & \textcolor{blue}{24.45} \\
    EEMFlow$_{\rm{ADM}}$ (Ours) & \textcolor{blue}{-72\%} & \textcolor{blue}{-83\%} & \textcolor{red}{12.75} & \textcolor{red}{33.87} & \textcolor{red}{10.64} & \textcolor{red}{32.44} & \textcolor{red}{22.43} \\
    \hline
    \end{tabular}%
  \label{tab:emd}%
\end{table*}

\section{Experiments}
\subsection{Implementation Details}
\subsubsection{Datasets}
\noindent{\bf HREM:} Our HREM dataset includes 100 indoor and outdoor scenes, with a resolution of 1280 $\times$ 720. We randomly select 70 scenes for training and reserve the remaining 30 for testing. The training set comprises $20,000$ samples, while the test set contains $8,000$ samples. Additionally, we further divide the test set by scene type (outdoor vs. indoor) and camera motion speed during rendering, resulting in four sub-sequences (outdoor\_slow, outdoor\_fast, indoor\_slow, indoor\_fast), with mean motion magnitudes ranging from, $0-30$, $30-100$, $0-20$ and $20-100$ pixels, respectively. Moreover, similar to~\cite{zhu2018ev}, we employ two data input settings: $dt=1$ and $dt=4$. $dt=1$ uses the event sequence between two consecutive frames of RGB images as input, with a meshflow generation frequency of 60 Hz, while $dt=4$ uses the event sequence between 4 consecutive frames of RGB images as input, with a meshflow generation frequency of 15 Hz.

\noindent{\bf HREM+:} We extend the HREM dataset to create a multi-density event dataset, named HREM+. Each sample in the training set consists of event sequences with varying densities, generated by adjusting the threshold $C$. The density range of the dataset spans from [0.05, 0.95]. \revised{We divide the data into multiple training sets based on different density ranges, ensuring that each training set comes from the same scene. Experiments show that the optimal density range for training is between [0.45, 0.55], as models trained on this range exhibit the best accuracy and generalization.} Each sample in the training set is paired with event data at this optimal density, which is used to supervise the training of the Adaptive Density Module (ADM), enabling it to effectively learn how to adjust input event data.

\noindent{\bf MVSEC:} The MVSEC dataset \cite{zhu2018multivehicle} is a real-world dataset collected in indoor and outdoor scenarios with sparse optical flow labels. As a common setting, $28,542$ data pairs of the `outdoor day2' sequence are used as the train set, and $8,410$ data pairs of the other sequences are used as the validation set. \revised{Additionally, based on Sec.~\ref{sec:meshflow_form_flow}, we convert the optical flow labels in MVSEC to meshflow, which is used for the domain generalization experiments from our HREM+ to real-world datasets with meshflow labels.}

\noindent{\bf DSEC:} The DSEC dataset \cite{gehrig2021dsec} is also collected using actual event cameras and lidar sensors on outdoor scenes. The dataset consists of $7,800$ training samples and $2,100$ test samples, divided into 24 sequences that include both day and night scenarios.

\subsubsection{Training details}
We conduct experiments using the PyTorch framework on two NVIDIA 2080Ti GPUs. We train all networks with the same parameters on our HRDM dataset. We employ the AdamW optimizer and OneCycle policy with a learning rate of $5 \times 10^{-4}$, weight decay of $5 \times 10^{-5}$, and other default parameters set to $\beta_{1} = 0.9, \beta_{2} = 0.99, \epsilon = 1 \times 10^{-4}$. 
For networks incorporating our ADM module, we first load the pre-trained weights and then jointly train the network and ADM on the extended HREM+ training set, using a learning rate of $1 \times 10^{-4}$.

\subsubsection{Evaluation metrics}
Following EV-FlowNet~\cite{zhu2018ev}, we use the average End-point Error (EPE) as the metric. When evaluating meshflow, both the prediction and the ground truth are upsampled to the input resolution to compute the metrics. Additionally, the DSEC dataset uses $N$PE to measure the percentage of flow errors greater than $N$ pixels in magnitude (e.g., 3PE, 2PE, 1PE) for flow outlier analysis, and employs Angular Error (AE) to assess directional accuracy. Besides EPE, MVSEC also uses \%Out, which represents the percentage of points with EPE greater than 3 pixels or 5\% of the ground truth flow magnitude.


\begin{figure*}[t]
    \centering
    \includegraphics[width=1.0\linewidth]{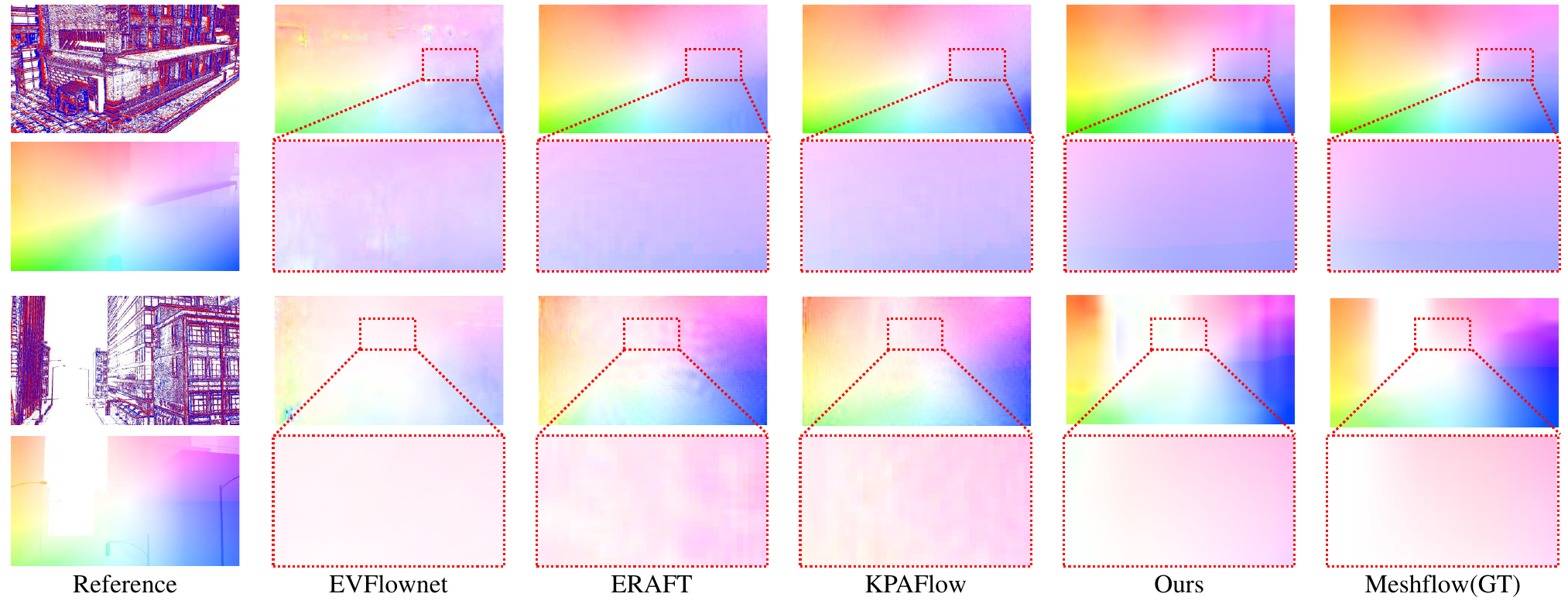}
    \vspace{-2em}
    \caption{Qualitative comparison of our proposed EEMFlow with other advanced flow networks on our HREM dataset. The subjective images of events and dense optical flow on the left side serve as references. The areas enclosed by red rectangles are zoomed in.}
    \label{fig:comparison}
\end{figure*}

\subsection{Comparison with State-of-the-Arts} 
\subsubsection{Results for Event-based Meshflow Estimation} 

\revised{In Tab.~\ref{tab:emd}, we train and test our EEMFlow on the HREM dataset, comparing it with several advanced networks, including EV-FlowNet\cite{zhu2019unsupervised}, ERAFT~\cite{gehrig2021raft}, PWCNet~\cite{sun2018pwc}, SKFlow~\cite{sun2022skflow}, GMA~\cite{jiang2021gma}, KPAFlow~\cite{luo2022kpa}, FlowFormer~\cite{huang2022flowformer}, DPFlow~\cite{morimitsu2025dpflow}, HDRFlow~\cite{Xu2024HDRFlow} for HDR imaging, and MeshHomoGan~\cite{liu2025meshhomogan}, DeepMeshflow~\cite{liu2025deepmeshflow} for video stabilization. Since pixel-based methods output predictions at the same resolution as the input, we upsample the ground truth $MF_{GT}$ during training to supervise these networks. However, mesh-based methods such as MeshHomoGan, DeepMeshflow, and our EEMFlow can directly output predictions at the same resolution as $MF_{GT}$. }For fair comparison, we upsample the predictions of all networks to a uniform resolution using bilinear interpolation during evaluation. We train and test all networks using two input settings ($dt=1$ and $dt=4$), and present their performance metrics across four test sub-sequences: outdoor\_slow, outdoor\_fast, indoor\_slow, indoor\_fast. Additionally, we report the average metric scores across all four test sub-sequences in the ``Avg" column.

\begin{table}[t]
  \centering
\caption{\revised{Comparison of image-based and event-based optical flow estimation methods on the DSEC benchmark. For image-based methods, consecutive RGB frames are used as input, while event-based methods use event sequences between the same frames. Event-based methods generally achieve superior performance in adverse conditions, highlighting the advantage of event data. $_{\rm{ADM}}$ indicates the use of the ADM plugin.}}
    \begin{tabular}
    {   >{\arraybackslash}p{2.7cm}| 
        >{\centering\arraybackslash}p{0.5cm}|
        >{\centering\arraybackslash}p{0.5cm}|
        >{\centering\arraybackslash}p{0.5cm}|
        >{\centering\arraybackslash}p{0.5cm}|
        >{\centering\arraybackslash}p{0.5cm}|
        >{\centering\arraybackslash}p{0.5cm}
    }
    \hline
    Methods & FPS$\uparrow$ & 1PE$\downarrow$ & 2PE$\downarrow$ & 3PE$\downarrow$ & EPE$\downarrow$ & AE$\downarrow$ \\
    \hline
    \multicolumn{7}{c}{\textit{Input Image Pair}} \\
    \hline
    RAFT~\cite{teed2020raft} & 12.3 & 12.4 & 4.60 & 2.61 & 0.78  & \textcolor{blue}{2.44} \\
    GMA~\cite{jiang2021gma} & 10.2 & 13.0 & 5.08 & 2.96 & 0.94 & 2.66 \\
    FlowDiffuser~\cite{luo2024flowdiffuser} & 6.29  & 12.2 & 4.82 & 2.77 &  0.82 & {2.59}\\
    DPFlow~\cite{morimitsu2025dpflow} & 9.92 & 12.1 & 4.48 & 2.56 & 0.78 & \textcolor{red}{2.41}\\
    \hline
    \multicolumn{7}{c}{\textit{Input Event Sequences}} \\
    \hline
    RTEF~\cite{brebion2021real} &- & 82.8 & 57.9 & 42.0 & 4.88 & 10.8 \\
    MutilCM~\cite{shiba2024secrets} &- & 76.6 & 48.5 & 30.9 & 3.47  & 14.0 \\
    MotionPriorCM~\cite{Hamann24motion} &- & 53.0 & 25.8 & 15.21 & 3.20  & 8.53 \\
    TamingCM~\cite{paredes2023taming} &- & 68.3 &34.2 &19.2 &2.49 &6.88 \\
    E2FAI~\cite{Guo25iccv} &\textcolor{red}{66.1}  & 37.5 & 17.6 & 11.2 & 1.78  & 6.44 \\
    OF-EV-SNN~\cite{cuadrado2023optical} &-  & 53.7 & 20.2 & 10.3 & 1.71  & 6.34 \\
    EVA-Flow~\cite{ye2023towards} &- &15.9 &- &3.20 & 0.88 & 3.31 \\
    ERAFT~\cite{gehrig2021raft} &11.4 & 12.7 & 4.74 & 2.68 & 0.79 & 2.85 \\
    ADMFlow~\cite{luo2023learning} &9.88 & {12.5} & {4.67} & {2.65} & {0.78} & {2.84} \\
    EFlowformer~\cite{li2023blinkflow} &- & {11.2} & {4.10} & {2.45} & {0.76} & {2.68} \\
    TMA~\cite{liu2023tma} &7.55 & {10.9} & {3.97} & {2.30} & {0.74} & {2.68} \\
    IDNet~\cite{wu2024lightweight} &12.8 & \textcolor{blue}{10.1} & {3.50} & {2.04} & {0.72} & {2.72} \\
    BFlow~\cite{Gehrig2024BFlow} & 13.1  & 11.9 & 4.41 & 2.44 & 0.75 & 2.68 \\
    EEMFlow+(Ours) &\textcolor{blue}{39.2} & 11.4 & {3.93} & {2.15} & {0.75} & {2.67} \\
    TMA$_{\rm{ADM}}$ & 6.72 & \textcolor{red}{9.97} & \textcolor{red}{3.48} & \textcolor{blue}{2.01} & \textcolor{red}{0.72} & 2.65 \\
    BFlow$_{\rm{ADM}}$ & 10.8 & {10.2} & \textcolor{blue}{3.49} & \textcolor{red}{1.99} & 0.72 & 2.65 \\
    EEMFlow+$_{\rm{ADM}}$(Ours) & 22.9 & 10.3 & 3.68 & 2.12 &\textcolor{blue}{0.73} & 2.70 \\
    \hline
    \end{tabular}
  \label{tab:dsec}
\end{table}

\begin{figure*}[t]
\centering
\includegraphics[width=\linewidth]{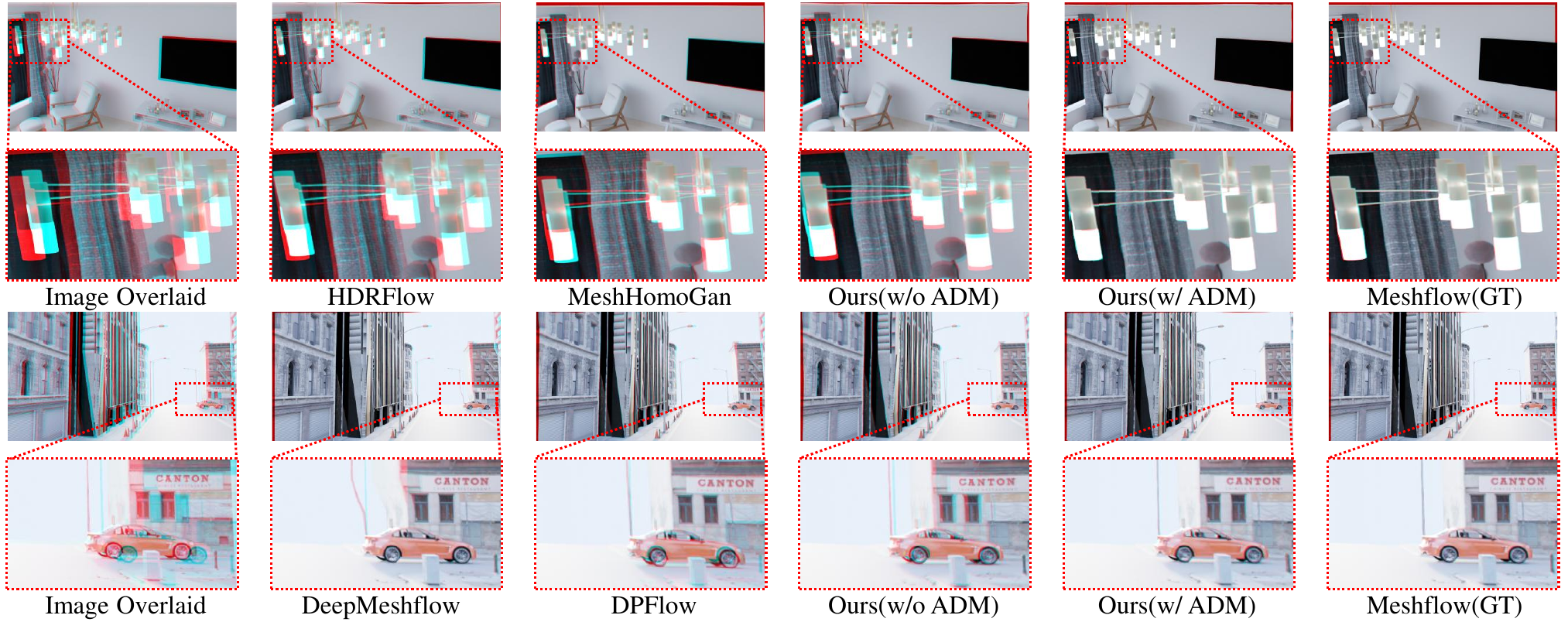}
\caption{\revised{Qualitative comparison of image alignment using estimated meshflow on the HREM+ dataset. We estimate meshflow from the event sequences and then warp image $I_{t_1}$ onto image $I_{t_2}$ for fusion, showcasing the fused result. The fewer blue or red ghosting artifacts indicate better alignment performance.}}
\label{fig:image_align_adm}
\end{figure*}

Tab.~\ref{tab:emd} shows that our EEMFlow achieves the lowest EPE score for both input settings $dt=1$ and $dt=4$ in the ``outdoor\_fast'' and ``indoor\_fast'' test sub-sequences, demonstrating its superior performance on high-speed and large-movement sequences. EEMFlow also exhibits great potential to outperform other flow networks on the ``outdoor\_slow'' and ``indoor\_slow'' test sub-sequences. Notably, EEMFlow achieves the lowest EPE scores in the ``Avg'' column for both input settings $dt=1$ and $dt=4$. Besides, we also achieves the least number of parameters and the fastest inference speed. Compared to ERAFT~\cite{gehrig2021raft}, our EEMFlow reduces the parameter count by 76\% (from 5.27 M to 1.24 M), reduces the inference time by 92\% (from 93 ms to 7 ms), and improves average EPE in $dt=4$ by 8\% (from 27.24 to 25.18). \revised{EEMFlow achieves comparable EPE performance to DPFlow~\cite{morimitsu2025dpflow} and FlowFormer~\cite{huang2022flowformer}, but is 30.3$\times$ and 38.7$\times$ faster in inference speed, respectively.} 

Additionally, to evaluate the impact of ADM, we conducted experiments by integrating ADM into EEMFlow to selectively adjust the densities of input events for meshflow estimation. The results, shown in Tab.~\ref{tab:emd}, demonstrate that the inclusion of ADM improves network performance for both $dt=1$ and $dt=4$ input settings. EPE decreased across all four sub-sequences of the HREM+ test set. Specifically, the average EPE across the four sub-sequences decreased by 8.18\% (from 5.50 to 5.05) for the $dt=1$ setting, and by 8.26\% (from 24.45 to 22.43) for the $dt=4$ setting.

In Fig.~\ref{fig:comparison}, we qualitatively compare our proposed EEMFlow with other flow networks, e.g., EV-FlowNet~\cite{zhu2019unsupervised}, ERAFT~\cite{gehrig2021raft}, and KPAFlow~\cite{luo2022kpa}. To facilitate comparison, we upsample the meshflow estimation and ground truth to the same resolution of input and zoom in on the areas with the apparent differences. EV-FlowNet shows the worst performance, with many holes and color mixing. ERAFT and KPAFlow would exhibit block artifacts and appear coarse in nature. In contrast, our EEMFlow results are smoother with more natural color transitions and greater similarity to the upsampled ground truth. 

\revised{In Fig.~\ref{fig:image_align_adm}, we present the subjective results of image alignment using the estimated meshflow predictions, including HDRFlow\cite{Xu2024HDRFlow}, MeshHomoGan~\cite{liu2025meshhomogan}, DeepMeshflow~\cite{liu2025deepmeshflow}, DPFlow~\cite{morimitsu2025dpflow}, and our EEMFlow (with and without ADM).} We estimate and upsample the meshflow prediction from the event sequences $E(t_1, t_2)$ and then warp image $I_{t_1}$ onto image $I_{t_2}$. The challenging alignment areas in the registered results are zoomed in, clearly demonstrating that our EEMFlow achieves excellent image alignment performance. The results show that EEMFlow with estimated meshflow provides outstanding alignment. Notably, after incorporating ADM, the image alignment performance of EEMFlow improves further, as evidenced by the reduced ghosting areas (indicated by the blue or red regions), which are the smallest among all methods.



\begin{figure*}[t]
\centering
\includegraphics[width=\linewidth]{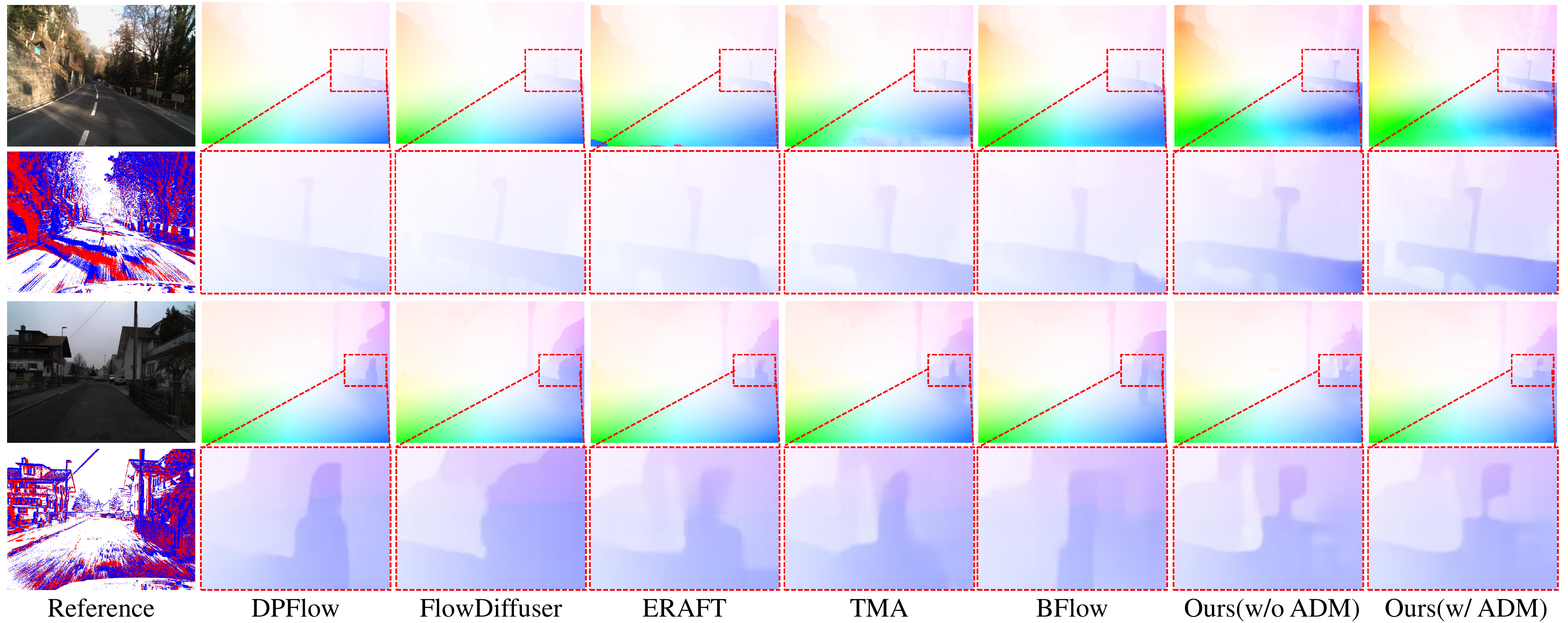}
\caption{\revised{Qualitative comparisons on the DSEC test set. We visualizethe dense predictions and zoom in the areas where are apparent differences.}}
\label{fig:comparisonDSEC_ADM}
\end{figure*}

\subsubsection{Results for Event-based Optical Flow Estimation} 
Tab.~\ref{tab:dsec} presents the comparative results on the DSEC dataset for our optical flow estimation network, EEMFlow+, against other event-based optical flow networks. EEMFlow+ achieves state-of-the-art performance in the 2PE, 3PE, and AE metrics, and performs on par with the best in the EPE metric. Notably, EEMFlow+ also maintains a significant advantage in inference speed. Compared to TMA\cite{liu2023tma}, EEMFlow+ increases the inference speed by 419\% (from 7.55 FPS to 39.2 FPS).
\revised{Additionally, we conduct experiments by integrating ADM into TMA~\cite{liu2023tma}, BFlow~\cite{liu2023tma}, and EEMFlow+ to selectively adjust the density of input events for optical flow estimation. We load the pre-trained weights of TMA, BFlow, and EEMFlow+ on the DSEC dataset, train them on the HREM+ dataset, and fine-tune them on DSEC with the ADM plugin.
Evaluation on the DSEC online benchmark shows that ADM further improves the performance of TMA, BFlow, and EEMFlow+, which already achieve high scores. Specifically, ADM reduces the 1PE of TMA by 8.26\% (from 10.9 to 10.0), BFlow by 14.29\% (from 11.9 to 10.2), and EEMFlow+ by 9.65\% (from 11.4 to 10.3). Fig.~\ref{fig:comparisonDSEC_ADM} demonstrates that ADM enables EEMFlow+ to produce optical flow estimates with sharper edges and more accurate shapes.} 

\subsection{Ablation Studies} 
\subsubsection{The Advantages of Event-Meshflow Estimation}
According to ERAFT~\cite{teed2020raft}, image-based approaches face challenges in handling difficult images due to the limited dynamic range of image sensors. \revised{We compare our methods with HDRFlow~\cite{Xu2024HDRFlow}, a flow method designed for HDR imaging, and the recent state-of-the-art optical flow methods DPFlow~\cite{morimitsu2025dpflow} and FlowFormer~\cite{huang2022flowformer}. For FlowFormer, HDRFlow, and DPFlow, we use consecutive image pairs as input for both optical flow and meshflow estimation. In contrast, EEMFlow+ and EEMFlow use event sequences as input for the same tasks. Image-based methods struggle with challenging images, such as overexposed or underexposed scenes, due to the limited dynamic range of image sensors. Even advanced image-based methods like DPFlow face difficulties in these scenarios. In comparison, our event-based methods (EEMFlow and EEMFlow+) achieve significantly better results. These conclusions are evident in Tab.~\ref{tab:advantages}, where EEMFlow achieves the lowest EPE scores on the HREM+ dataset, highlighting its efficiency and suitability for real-time applications such as online video stabilization and autonomous driving.}

\begin{table}[t]
  \centering
  \caption{\revised{Results of meshflow and optical flow on HREM+ dataset for $dt=1$.}}
    \begin{tabular}
    {   >{\arraybackslash}p{0.8cm}| 
        >{\centering\arraybackslash}p{2.0cm}|
        >{\centering\arraybackslash}p{0.6cm}
        >{\centering\arraybackslash}p{0.6cm}|
        >{\centering\arraybackslash}p{0.6cm}
        >{\centering\arraybackslash}p{0.6cm}|
        >{\centering\arraybackslash}p{0.6cm}
    }
    \hline
    \multirow{2}{*}{Task} & \multirow{2}{*}{Method}  & \multicolumn{2}{c|}{Outdoor} & \multicolumn{2}{c|}{Indoor} & \multirow{2}{*}{Avg}\\
    & & Slow & Fast & Slow & Fast & \\
    \hline
            & FlowFormer~\cite{huang2022flowformer} &6.20 &16.06 &5.99 &15.27 & 10.88 \\
    Optical & HDRFlow~\cite{Xu2024HDRFlow} &5.98 & 16.21 & 5.84 & 15.68 & 10.93 \\
    Flow & DPFlow~\cite{morimitsu2025dpflow} &4.13 & 13.94 & 4.15 & 13.44 & 8.92  \\
         & EEMFlow+ &3.88 &11.02 &4.03 &10.92  &7.46 \\
    \hline
        & FlowFormer~\cite{huang2022flowformer} &5.99 &15.12 &5.74 &14.95 &10.45 \\    
    Mesh- & HDRFlow~\cite{Xu2024HDRFlow} &5.87 & 14.84 & 4.92 & 13.92 & 9.89  \\
    flow & DPFlow~\cite{morimitsu2025dpflow} & 3.95 & 13.08 & 4.01 & 13.37 & 8.60  \\
        & EEMFlow &\textcolor{red}{2.42} &\textcolor{red}{9.09}  &\textcolor{red}{2.00}  &\textcolor{red}{8.46}  &\textcolor{red}{5.50} \\
    \hline
    \end{tabular}
  \label{tab:advantages}
\end{table}

\begin{table}[t]
  \centering
  \caption{\revised{Ablation studies about CDC of EEMFlow+ on DSEC dataset. CDC consists of two branches, the self-corrector and self-correlation.}}
    \begin{tabular}
    {   >{\arraybackslash}p{1.6cm}|
        >{\centering\arraybackslash}p{0.9cm}|
        >{\centering\arraybackslash}p{1.1cm}|
        >{\centering\arraybackslash}p{0.5cm}|
        >{\centering\arraybackslash}p{0.5cm}|
        >{\centering\arraybackslash}p{0.5cm}|
        >{\centering\arraybackslash}p{0.5cm}
    }
    \hline
    \multirow{2}{*}{Model} & Self- & Self- & \multirow{2}{*}{FPS$\uparrow$}  & \multirow{2}{*}{1PE$\downarrow$}  & \multirow{2}{*}{EPE$\downarrow$}  & \multirow{2}{*}{AE$\downarrow$} \\
    & corrector &correlation & & & &\\
    \hline
    EEMFlow   &  \ding{56} &  \ding{56}  & \textcolor{red}{161.3}  & 18.3  & 1.02  & 4.38 \\
    \hline
    (a)EEMFlow+   &  \ding{56} &  \ding{56}  & {60.4}  & 15.5  & 0.89  & 3.11 \\
    (b)EEMFlow+   &  \ding{52} &  \ding{56}  & 55.6  & 14.1  & 0.81  & 2.92 \\
    (c)EEMFlow+   &  \ding{56} &  \ding{52}  & 46.3  & 12.8  & 0.79  & 2.78 \\
    (d)EEMFlow+   &  \ding{52} &  \ding{52}  & 39.2  & \textcolor{red}{11.4}  & \textcolor{red}{0.75}  & \textcolor{red}{2.67} \\
    \hline
    \end{tabular}
  \label{tab:ablation_CDC}
\end{table}

\begin{table}[t]
    \centering
        \caption{\revised{Domain generalization experiment on MVSEC dataset for event-based meshflow estimation. ``M'' denotes training on MVSEC only, while ``H+M'' indicates pretraining on HREM+ followed by fine-tuning on MVSEC.}}
    \resizebox*{\linewidth}{!}{
        \begin{tabular}
            {   >{\arraybackslash}p{2.5cm}| 
                >{\centering\arraybackslash}p{0.8cm}| 
                >{\centering\arraybackslash}p{0.8cm} 
                >{\centering\arraybackslash}p{0.8cm}| 
                >{\centering\arraybackslash}p{0.8cm} 
                >{\centering\arraybackslash}p{0.8cm} 
            }
\hline
\multirow{2}{*}{Method } & Train& \multicolumn{2}{c|}{$dt=1$} & \multicolumn{2}{c}{$dt=4$}  \\
&D.Set&EPE$\downarrow$ & \%Out$\downarrow$ &EPE$\downarrow$ & \%Out$\downarrow$\\
    \hline
    \multirow{2}{*}{HDRFlow~\cite{Xu2024HDRFlow}}  & M     & 0.92  & 2.84     & 3.54  & 41.26  \\
              & H+M  & 0.64  & 0.58   & 2.31  & 24.11  \\
    \hline
    \multirow{2}{*}{MeshHomoGan~\cite{liu2025meshhomogan}}  & M     & 0.99  & 3.07    & 3.61  & 42.31  \\
              & H+M  & 0.75  & 1.56   & 2.24  & 24.42  \\
    \hline
    \multirow{2}{*}{DeepMeshflow~\cite{liu2025deepmeshflow}} & M     & 0.87  & 2.38    & 3.37  & 39.88  \\
     & H+M  & 0.59  & 0.42    & 2.01  & 23.85  \\
    \hline
    \multirow{2}{*}{DPFlow~\cite{morimitsu2025dpflow}}& M     & 0.82  & 1.94      & 2.92  & 34.17  \\
              & H+M  & \textcolor{blue}{0.57}  &\textcolor{blue}{0.33}  & 1.91  & \textcolor{blue}{19.84}  \\
    \hline
    \multirow{2}{*}{EEMFlow(Ours)}& M     & 0.88  & 2.02    & 2.89  & 33.80  \\
              & H+M  & {0.58}  & 0.39  & \textcolor{blue}{1.90} & {20.10} \\
    \hline
    \multirow{2}{*}{EEMFlow$_{\rm{ADM}}$(Ours)}& M     & -  & -    & -  & -  \\
              & H+M   & \textcolor{red}{0.52} & \textcolor{red}{0.29} & \textcolor{red}{1.82}  & \textcolor{red}{15.80}  \\
    \hline
    \end{tabular}}
    \label{tab:HREM+_MVSEC_domin}
\end{table}

\begin{figure*}[t]
\centering
\includegraphics[width=\linewidth]{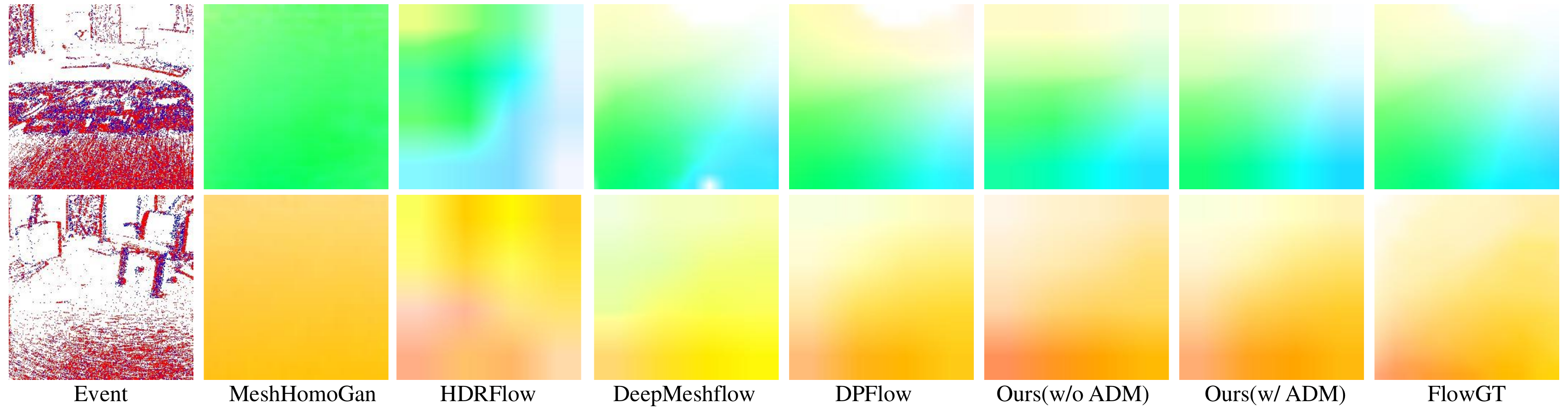}
\caption{\revised{Qualitative comparison of meshflow estimation results on MVSEC test sequences. The visualizations show two representative samples from the dataset, where networks are trained on the HREM+ and MVSEC training set.}}
\label{fig:MVSEC_meshflow_compair}
\end{figure*}

\subsubsection{Experiments for CDC of EEMFlow+}
In Tab.~\ref{tab:ablation_CDC}, we also conduct ablation experiments on the CDC module of EEMFlow+ used for optical flow estimation, including its two branches, the self-corrector and self-correlation. We train and test all models using the same settings on the DSEC dataset to show the individual impact of each branch in CDC module. Comparison of (a)\&(b) demonstrates that the CDC with only the self-corrector branch can bring a significant increase in accuracy with a minimal loss in speed. Comparison of (b)\&(c) shows that self-correlation, compared to self-corrector, can lead to a higher increase in accuracy, albeit at a further reduction in inference speed. Finally, the comparison of (a)\&(d) shows that the CDC composed of both the self-corrector and self-correlation branches significantly improves the accuracy of optical flow estimation with an acceptable loss in speed.

\section{Extended Experiments}

\subsection{\revised{Domain Generalization of HREM+ Dataset.}}
\revised{
To evaluate domain generalization capabilities, we conduct experiments on the MVSEC dataset for event-based meshflow estimation. To our knowledge, no existing real event camera dataset provides meshflow labels. Following the Motion Propagation and Median Filters process described in Sec. \ref{sec:meshflow_form_flow}, we convert the optical flow labels from the MVSEC dataset into meshflow labels. 
As shown in Tab.~\ref{tab:HREM+_MVSEC_domin}, we evaluate several advanced networks (HDRFlow~\cite{Xu2024HDRFlow}, MeshHomoGAN~\cite{liu2025meshhomogan}, DeepMeshflow~\cite{liu2025deepmeshflow}, DPFlow~\cite{morimitsu2025dpflow}, EEMFlow without and with ADM) under two training settings: (1) trained solely on the MVSEC training set (M), and (2) pre-trained on the synthetic HREM+ dataset and then fine-tuned on MVSEC (denoted as H+M). All models are tested on four MVSEC test sequences: indoor\_flying1, indoor\_flying2, indoor\_flying3, and outdoor\_day1. We report the average EPE and \%Out.}

\begin{figure}[t]
    \centering
    \includegraphics[width=\linewidth]{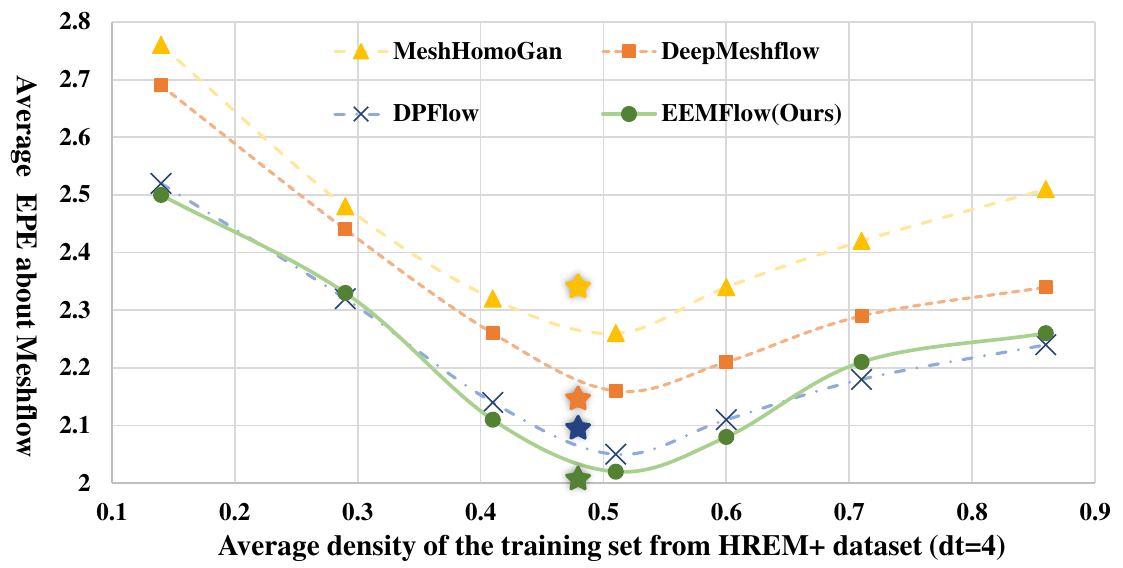}
    \caption{\revised{Effect of training set density on event-based meshflow estimation. The input is set to $dt=4$ (corresponding to the temporal window spanning four consecutive RGB camera exposures), and the models are evaluated on MVSEC (meshflow labels are converted from optical flow labels). The pentagons highlight the performance of networks trained on the complete HREM+ dataset, with densities ranging from [0.05, 0.95] and an average density of 0.48.}}
    \label{fig:different_density_meshflow}
\end{figure}

\begin{figure}[t]
    \centering
    \includegraphics[width=\linewidth]{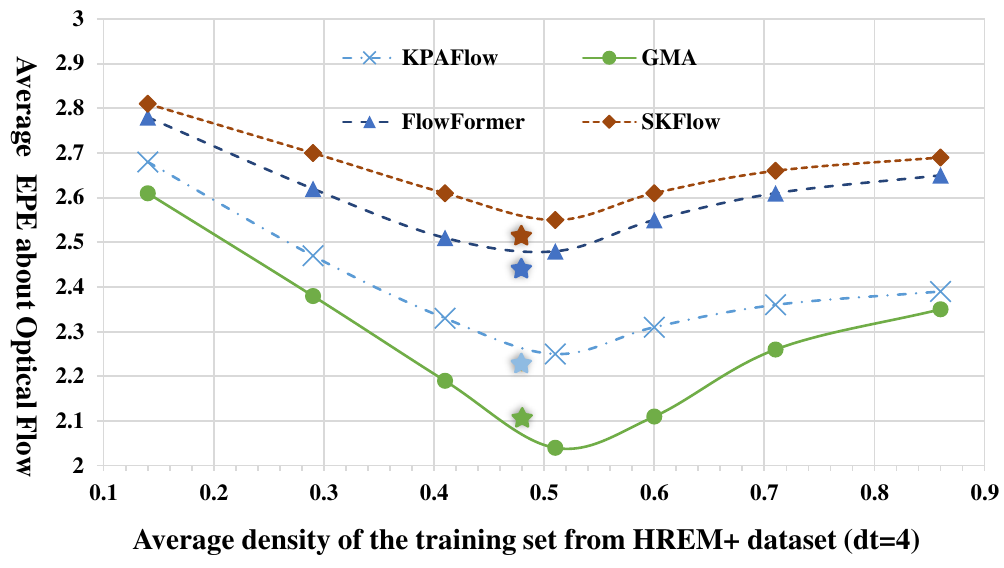}
    \caption{\revised{Effect of training set density on event-based optical flow estimation. The input is set to $dt=4$ (corresponding to the temporal window spanning four consecutive RGB camera exposures), and the models are evaluated on the real event dataset MVSEC with fixed event density. The pentagons highlight the performance of networks trained on the complete HREM+ dataset, with densities ranging from [0.05, 0.95] and an average density of 0.48.}}
    \label{fig:different_density_optical_flow}
\end{figure}

\revised{
The results demonstrate that all networks trained on the H+M training set outperform those trained only on MVSEC (M), indicating that the HREM+ dataset can transfer to real event sequences. Fig.~\ref{fig:MVSEC_meshflow_compair} shows two visual examples of the test results of the networks trained on the H+M training set. Notably, our EEMFlow with ADM achieves state-of-the-art performance, demonstrating superior accuracy in real-world event-based meshflow estimation tasks.}

\subsection{Comparison of Different Event Densities}
\subsubsection{\revised{Comparison Across Different Densities on HREM+}} \label{densities_ablation_HREM+}
\revised{
We configure the event data density range for the HREM+ training set to [0.05, 0.95], with an average density of 0.48. To investigate the impact of density on flow estimation performance, we partition the HREM+ training set into seven density ranges: [0.05, 0.25], [0.25, 0.35], [0.35, 0.45], [0.45, 0.55], [0.55, 0.65], [0.65, 0.75], and [0.75, 0.95]. The corresponding average densities for these subsets are 0.14, 0.29, 0.41, 0.51, 0.60, 0.71, and 0.86, respectively. We then train networks on each subset independently and evaluate their performance on real event data from the MVSEC dataset.
For event-based meshflow estimation, we employ MeshHomoGan \cite{liu2025meshhomogan}, DeepMeshflow \cite{liu2025deepmeshflow}, and our proposed EEMFlow \cite{Luo_2024_EEMFlow} for mesh-level flow estimation, along with DPFlow \cite{morimitsu2025dpflow} for pixel-level meshflow estimation. For event-based optical flow estimation, we utilize SKFlow \cite{sun2022skflow}, GMA \cite{jiang2021gma}, FlowFormer \cite{huang2022flowformer}, and KPA-Flow \cite{luo2022kpa}.
}

\revised{
Fig.~\ref{fig:different_density_meshflow} presents the results for event-based meshflow estimation using an input setting of $dt=4$ (corresponding to the temporal window spanning four consecutive RGB camera exposures). Networks are trained on event data within different density ranges and subsequently tested on the MVSEC dataset with fixed density. Similarly, Fig.~\ref{fig:different_density_optical_flow} illustrates the results for event-based optical flow estimation under the same input setting $dt=4$.}
\revised{
The experimental results demonstrate that model performance improves progressively as the average density of the training set increases. However, performance begins to deteriorate when the average density becomes excessively high. Notably, the density range [0.45, 0.55] consistently achieves optimal performance for both meshflow and optical flow estimation tasks, suggesting the existence of an optimal density regime for training event-based flow estimation networks.
}

\begin{figure}[t]
    \centering
    \includegraphics[width=\linewidth]{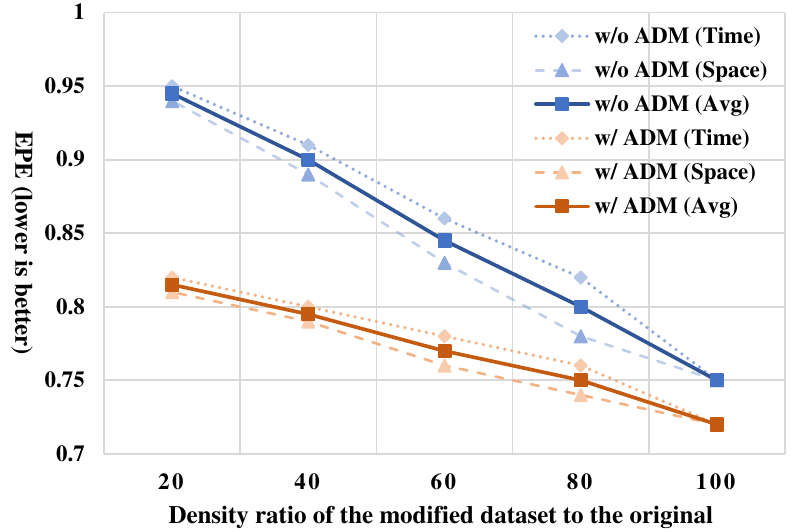}
    \vspace{-.5cm}
    \caption{\revised{Comparison of training on DSEC dataset with varying densities. Event data with different densities are generated using Spatial-guided (Space) and Temporal-guided (Time) sampling strategies from the DSEC train set. EEMFlow+ (with and without ADM) is trained on these subsets and tested on the original DSEC test set. "Avg" represents the combined performance of both strategies.}}
    \label{fig:density_dsec}
\end{figure}

\subsubsection{\revised{Comparison Across Different Densities on DSEC}} \label{densities_ablation_DSEC}
\revised{
We conduct analogous experiments on the DSEC dataset to evaluate the impact of density variations on real-world event data. The DSEC dataset exhibits fixed event data density, with an average density of 0.41 for the training set. As shown in Fig.~\ref{fig:density_dsec}, two physics-compliant subsampling methods guided by optical flow are considered: spatial-guided sampling and temporal-guided sampling, applied separately.}

\revised{
For spatial-guided sampling, we uniformly sample a subset of spatial locations across the image plane. Utilizing ground-truth optical flow $F_{i \to j}$, we trace the motion trajectories of these sampled pixels through time and retain only events that lie along these warped paths. This approach preserves local event motion patterns while avoiding arbitrary fragmentation of edge contours.}
\revised{
For temporal-guided sampling, we subsample a subset of active timestamps and retain events occurring at those specific times. To maintain consistency, we restrict the retained events to those that are spatially consistent with optical flow trajectories at the sampled timestamps. These strategies ensure that the downsampled events approximate motion trajectories consistent with the underlying scene dynamics and mitigate correlation loss due to random removal.}

\begin{figure*}[t]
    \centering
    \includegraphics[width=\linewidth]{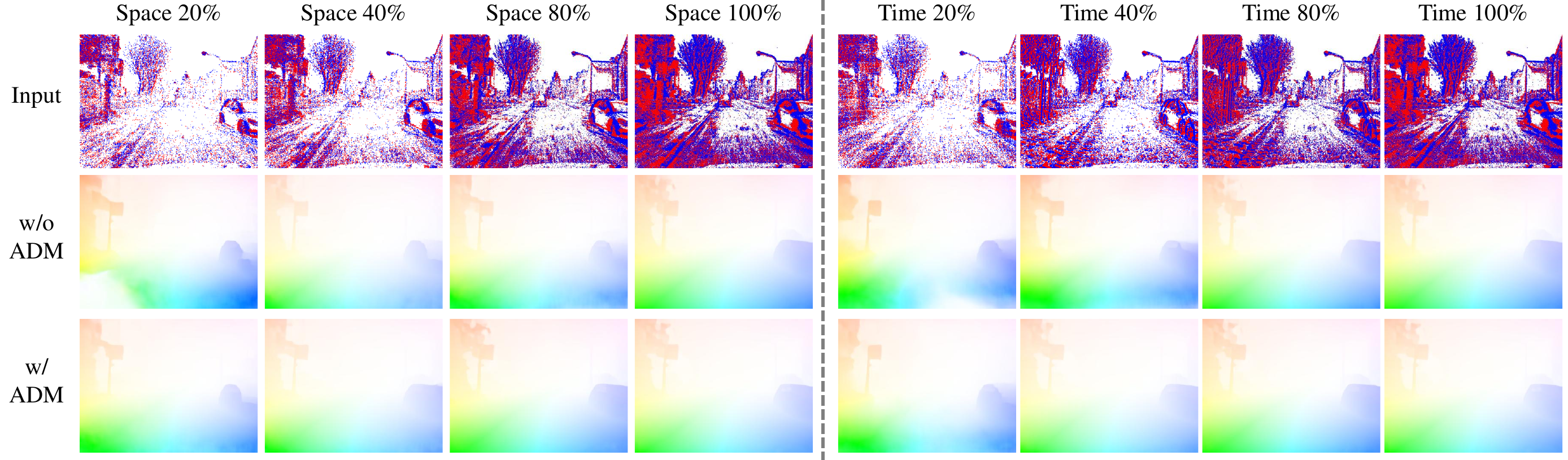}
    \caption{\revised{Qualitative Comparison of Optical Flow Estimation for Varying Event Data Densities. Event data sampled at 20\%, 40\%, 60\%, 80\%, and 100\% densities using Spatial-guided (Space) and Temporal-guided (Time) sampling are processed by EEMFlow+ (with and without ADM) for the same scene.}}
    \label{fig:DSEC_multi_density}
\end{figure*}

\revised{We apply each sampling method to the DSEC training set with density ratios of 20\%, 40\%, 60\%, 80\%, and 100\%, preserving spatio-temporal consistency. As the sampling ratio decreases, the corresponding event data density also decreases proportionally. We utilize both sampling strategies to generate event training data at different densities and train our EEMFlow+ with and without ADM, subsequently evaluating these models on the original DSEC test set.
Fig.~\ref{fig:density_dsec} illustrates the EPE trends across different sampling ratios, demonstrating the accuracy of optical flow estimation. Fig.~\ref{fig:DSEC_multi_density} provides a qualitative comparison of optical flow estimates from our EEMFlow+ (both with and without ADM) for event data sampled at 20\%, 40\%, 60\%, 80\%, and 100\% ratios. The results indicate that optical flow estimation error increases progressively as event data density decreases, regardless of whether spatial-guided or temporal-guided sampling is employed.} 

\begin{figure*}[t]
    \centering
    \includegraphics[width=\linewidth]{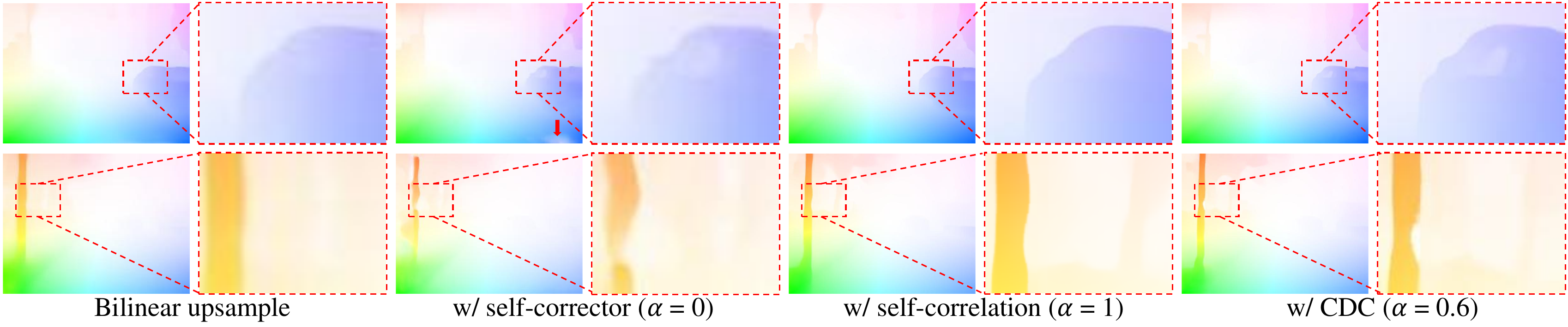}
    \caption{\revised{Abaltion study on the impact of $\alpha$ in the CDC module. Only the self-corrector branch is active when $\alpha = 0$, while only the self-correlation branch is active when $\alpha = 1$.}}
    \label{fig:CDC_alpha_ablation}
\end{figure*}

\subsection{Parameter Settings and Strategy Selection}
\begin{figure*}[t]
\centering
\includegraphics[width=\linewidth]{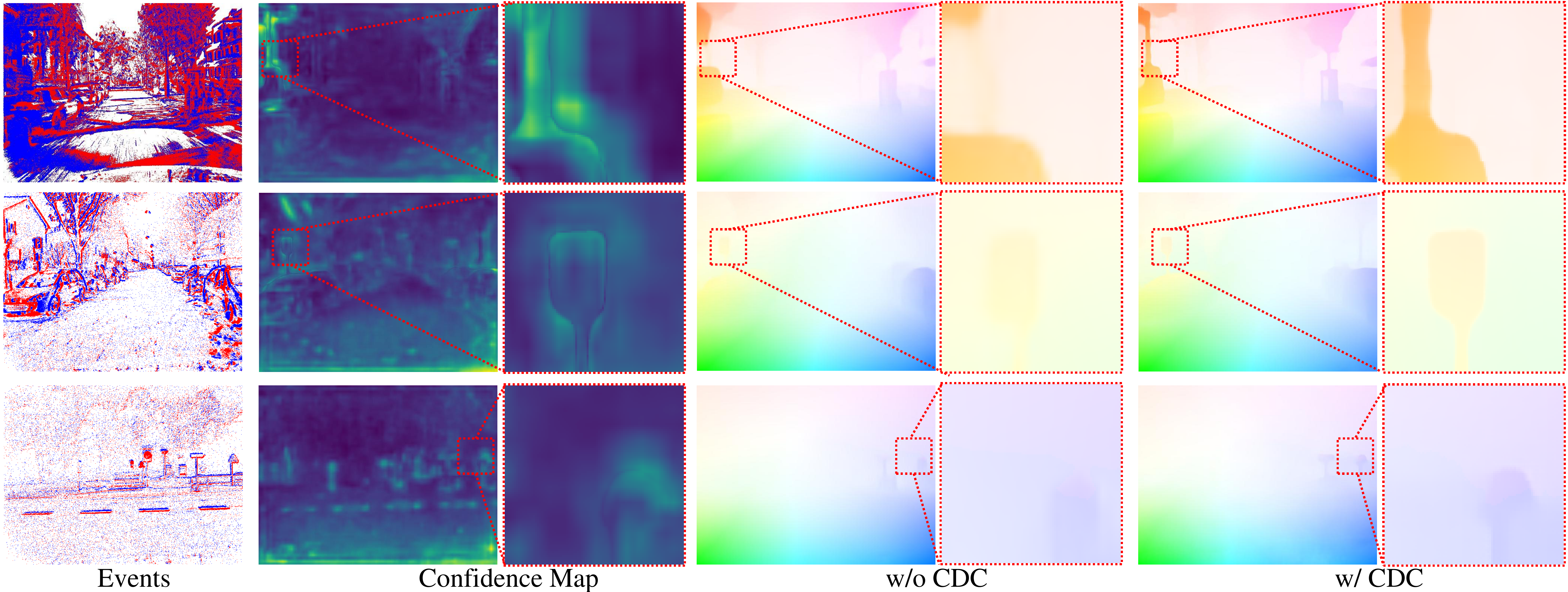}
\caption{\revised{Visualization of confidence maps under varying event data densities. Brighter colors indicate higher confidence. Optical flow predictions of EEMFlow+ with and without CDC are also shown for each input density.}}
\label{fig:confidence_map_accuracy}
\end{figure*}

\subsubsection{\revised{The Setting of Parameter $\alpha$ in CDC}}
\revised{$\alpha$ is a fixed hyperparameter in our Confidence-induced Detail Completion (CDC) module. The $\alpha$ parameter is used within the CDC module to balance the contributions of two branches: the self-corrector branch and the self-correlation branch. The self-corrector branch corrects blurred and expanded edges caused by bilinear upsampling, while the self-correlation branch, based on self-attention mechanisms, refines the motion in error regions, particularly around the edges. To determine the optimal value for $\alpha$, we conduct an ablation study. Initially, we find that $\alpha = 0.6$ yield the best performance in terms of improving edge clarity and reducing errors across a range of motion scenarios. As shown in Fig.~\ref{fig:CDC_alpha_ablation}, bilinear upsampling leads to blurry and expanded edges; the self-corrector branch can fix the expanded areas but does not fully restore sharpness, while the self-correlation branch can enhance edge clarity but cannot fix the expansion. By combining both branches, we can address both issues simultaneously.}

\subsubsection{\revised{The Accuracy of Confidence Map in CDC}}
\begin{figure}[t]
\centering
\includegraphics[width=\linewidth]{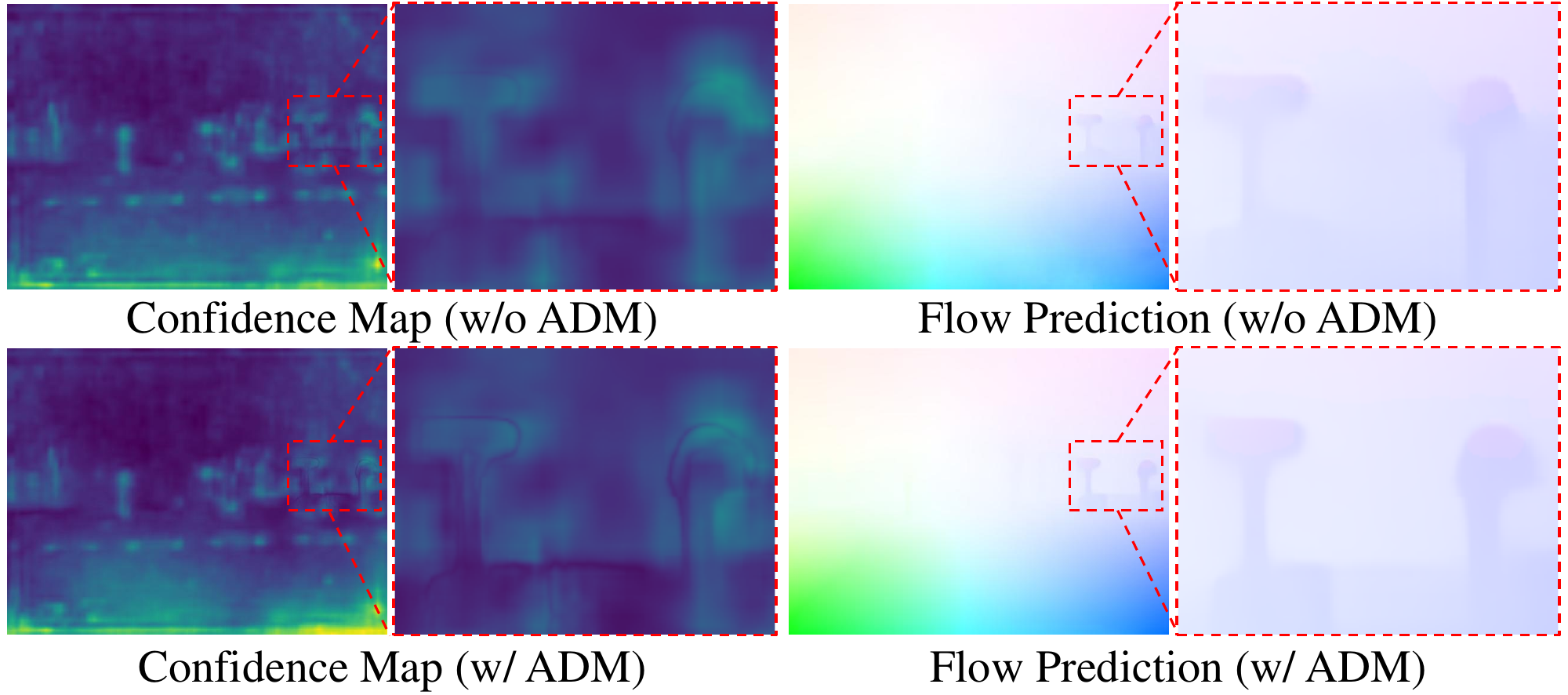}
\caption{\revised{Effect of ADM on confidence map accuracy. Comparison of confidence maps and flow predictions of EEMFlow+ without and with ADM.}}
\label{fig:confidence_map_accuracy_adm}
\end{figure}

\revised{The confidence map $W^i$ is trained in a learnable manner without direct ground truth supervision. Instead, it is indirectly supervised through the use of flow ground truth. We visualize the confidence maps under different event data densities (Our decoder consists of 5 layers, requiring 4 upsampling operations). For each upsampling stage, the confidence map $W^i$ is upsampled to the output size. We then compute a weighted sum of all $W^i$ maps using $2^i$ as the weight for each layer, followed by normalization to the range $[0,1]$. The resulting confidence map increases from 0 to 1, and the corresponding visualization shows a gradient from dark to light.
Fig.~\ref{fig:confidence_map_accuracy} demonstrates $W^i$ that effectively helps the CDC module restore and sharpen object boundaries by leveraging event signals at the object edges. However, the figure also shows that when the event data density decreases, the confidence map's ability to detect edge regions becomes less pronounced, leading to a decline in the ability to refine and clarify object edges for CDC.
To address this issue, we introduce the Adaptive Density Module (ADM), which enhances the input event data density. This helps the confidence map better detect edge regions and improves the ability of CDC to restore sharp object boundaries, ultimately leading to more accurate optical flow estimation. As shown in Fig.~\ref{fig:confidence_map_accuracy_adm}, the ADM allows the confidence map to detect more prominent object edges, thereby improving flow prediction accuracy, especially in edge regions. }

\begin{table}[t]
\centering
\caption{\revised{Ablation of channel shuffle in the EEMFlow decoder on HREM+. "Slow" denotes the average EPE of Indoor\_slow and Outdoor\_slow, while "Fast" denotes the average EPE of Indoor\_fast and Outdoor\_fast.}}
    \begin{tabular}    
    {   >{\arraybackslash}p{2.0cm}| 
        >{\centering\arraybackslash}p{0.8cm} 
        >{\centering\arraybackslash}p{0.8cm}| 
        >{\centering\arraybackslash}p{0.8cm} 
        >{\centering\arraybackslash}p{0.8cm}| 
        >{\centering\arraybackslash}p{1.0cm} 
    }
    \hline
    \multirow{2}{*}{Convolutions} & \multicolumn{2}{c|}{Avg($dt=1$)} & \multicolumn{2}{c|}{Avg($dt=4$)} & Param. \\
       & Slow   & Fast & Slow   & Fast & (M) \\
    \hline
    Standard & 2.59  & 9.45  & 13.94  & 38.34  & 3.06  \\
    Depthwise & 2.66  & 11.79  & 14.81  & 40.18  & 0.98  \\
    \underline{Channel Shuffle} & \textcolor{red}{2.21}  & \textcolor{red}{8.78}  & \textcolor{red}{13.03}  & \textcolor{red}{35.86}  & \textcolor{red}{1.24}  \\
    \hline
    \end{tabular}%
    \label{tab:ablation_shuffle_block}
\end{table}

\begin{table}[t]
\centering
\caption{\revised{Ablation of Dilated Feature Correlation (DFC) in EEMFlow on HREM+. }}
    \begin{tabular}    
    {   >{\arraybackslash}p{2.4cm}| 
        >{\centering\arraybackslash}p{0.8cm} 
        >{\centering\arraybackslash}p{0.8cm}| 
        >{\centering\arraybackslash}p{0.8cm} 
        >{\centering\arraybackslash}p{0.8cm}| 
        >{\centering\arraybackslash}p{1.0cm} 
    }
        \hline
        \multirow{2}{*}{Correlation Grid} & \multicolumn{2}{c|}{Avg($dt=1$)} & \multicolumn{2}{c|}{Avg($dt=4$)} & Time \\
         & Slow   & Fast & Slow   & Fast & (ms) \\
        \hline
        Standard($r=3$)   & 2.58  & 9.05  & 13.89  & 36.62  & 7  \\
        Standard($r=4$)   & {2.29}  & \textcolor{red}{8.72}  & {13.04}  & \textcolor{red}{35.69}  & {27}  \\
        \underline{DFC}   & \textcolor{red}{2.21}  & {8.78}  & \textcolor{red}{13.03}  & {35.86}  & \textcolor{red}{7}  \\
        \hline
        \end{tabular}%
        \label{tab:ablation_correlation}
\end{table}

\begin{figure*}[t]
\centering
\includegraphics[width=\linewidth]{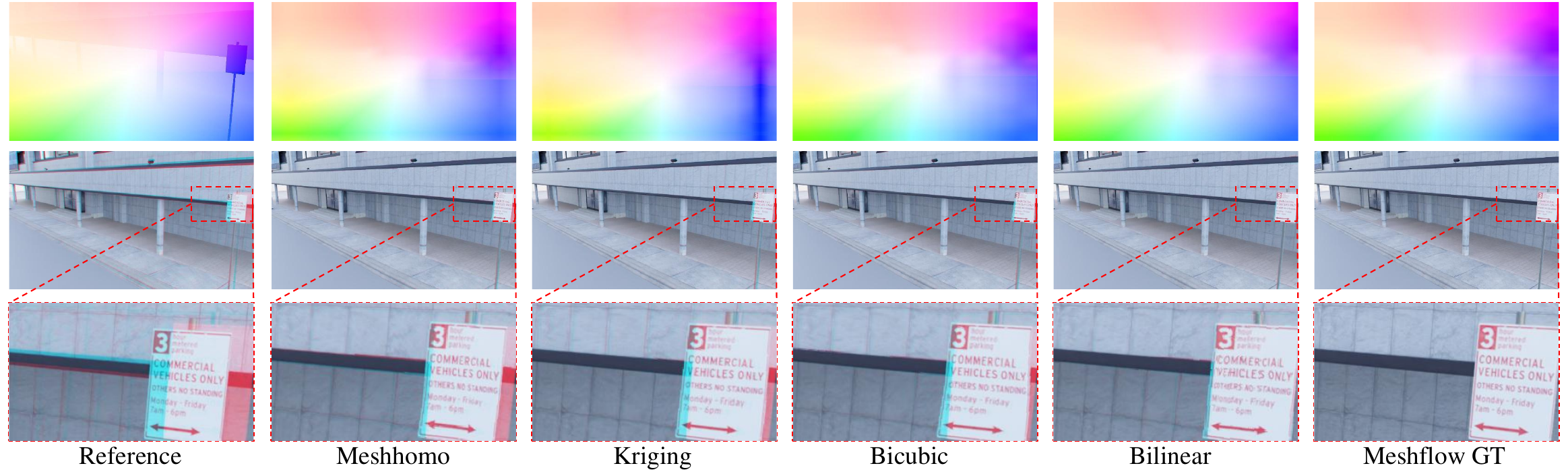}
\caption{\revised{Visualization of meshflow upsampling using different interpolation methods, with corresponding alignment results. The meshflow is upsampled from  EEMFlow prediction.}}
\label{fig:meshflow_interpolation}
\end{figure*}

\begin{table}[t]
\centering
\caption{\revised{Ablation of the ADM softmax weighting strategy in EEMFlow w/ ADM.}}
    \begin{tabular}    
    {   >{\arraybackslash}p{2.0cm}| 
        >{\centering\arraybackslash}p{0.8cm} 
        >{\centering\arraybackslash}p{0.8cm}| 
        >{\centering\arraybackslash}p{0.8cm} 
        >{\centering\arraybackslash}p{0.8cm}| 
        >{\centering\arraybackslash}p{1.0cm} 
    }
        \hline
        \multirow{2}{*}{Correlation Grid} & \multicolumn{2}{c|}{Avg($dt=1$)} & \multicolumn{2}{c|}{Avg($dt=4$)} & Param. \\
         & Slow   & Fast & Slow   & Fast & (M) \\
        \hline
        Only $V^{{\rm{MDC}}}$  & 2.09  & 8.66  & 11.98  & 33.99  & 1.46  \\
        Average fusion   & {2.18}  & {8.79}  & {12.54}  & {34.68}  & {1.46}  \\
        \underline{Softmax}   & \textcolor{red}{2.06}  & \textcolor{red}{8.05}  & \textcolor{red}{11.70}  & \textcolor{red}{33.16}  & \textcolor{red}{1.48}  \\
        \hline
        \end{tabular}%
        \label{tab:ablation_softmax}
\end{table}

\subsubsection{\revised{Channel Shuffle and DFC of EEMFlow}}
\revised{We conduct an ablation study on the Channel Shuffle blocks in the decoder of EEMFlow by replacing the three Channel Shuffle layers with either standard convolutions or depthwise separable convolutions. As shown in Tab.~\ref{tab:ablation_shuffle_block}, we evaluate the average EPE (lower is better) on the Slow setting (indoor\_slow and outdoor\_slow) and the Fast setting (indoor\_fast and outdoor\_fast) of HREM+. The results show that although depthwise separable convolutions reduce the network parameters, they lead to a significant performance drop, whereas Channel Shuffle achieves a better trade-off, maintaining high accuracy while keeping the parameter count low. 
We also perform an ablation on the Dilated Feature Correlation (DFC) in EEMFlow. In this experiment, we replace DFC with standard correlation using a square search range of size $(2r+1)\times(2r+1)$, with $r=3$ and $r=4$. 
Our DFC masks out part of the search grid while maintaining the same receptive field as $r=4$, but with the computational cost close to $r=3$. The results in Tab.~\ref{tab:ablation_correlation} confirm that DFC improves computational efficiency while preserving high accuracy.}

\subsubsection{\revised{Softmax Weighting Mechanism of ADM}}
\revised{
We conduct an ablation on the softmax weighting mechanism of ADM in EEMFlow w/ ADM, as shown in Tab.~\ref{tab:ablation_softmax}. The softmax operation produces pixel-wise weights for fusing the enhanced event representation $V^{{\rm{MDC}}}$ from the MDC module with the original $V$. We compare it against two alternatives: (i) using only $V^{{\rm{MDC}}}$ as the ADM output, and (ii) average fusion, where the output is the pixel-wise mean of $V^{{\rm{MDC}}}$ and $V$. The Results show that the softmax-based fusion achieves the best performance, as it adaptively selects the optimal pixel-wise density representation.}

\subsubsection{\revised{Upsampling Strategy for Meshflow}}
\revised{To evaluate the impact of different interpolation techniques on meshflow alignment, we compare bilinear interpolation with Meshhomo, Kriging, and Bicubic.
Meshhomo incorporates grid-aware constraints, as described in MeshHomoGan~\cite{liu2025meshhomogan}. In this approach, the displacement at each of the four vertices of a mesh grid is used to compute a 3$\times$3 homography matrix for that grid. This displacement is then applied to all pixels within the mesh grid. The interpolation smooths transitions between mesh grids, as 15$\times$15 mesh grid vertices are involved in the homography calculation for neighboring grids. Kriging interpolation is a more advanced technique that uses spatial covariance to build a spatial autocorrelation model. This model is then used to perform interpolation at new positions, enabling more accurate upsampling. Bilinear and Bicubic interpolation are the linear weighted average methods based on neighboring pixels.
}

\revised{
We train EEMFlow on the HREM+ test set and estimate meshflow displacements (from Image1 to Image2). We then apply the four interpolation methods and use backward warping to align Image2 to Image1. Fig.~\ref{fig:meshflow_interpolation} shows the upsampling results and image alignment for each method. The results show that bilinear interpolation extracts global motion information more effectively, reducing the interference from local displacements (e.g., large objects like the billboard on the right) and yielding better alignment performance. For meshflow, low-rank global
motion typically involves large-scale uniform changes, such as translation, rotation, and scaling. Bilinear interpolation is particularly well-suited for these types of scenarios.}

\section{Conclusion}
In this work, we develop the first event-based meshflow dataset, HREM, featuring 100 indoor and outdoor virtual scenes rendered with Blender, offering physically accurate event data and meshflow labels at the highest resolution to date. We propose the Efficient Event-based MeshFlow (EEMFlow) network, which achieves state-of-the-art performance on the HREM dataset while maintaining high efficiency. Building on EEMFlow, we introduce a Confidence-induced Detail Completion (CDC) module, upgrading it to EEMFlow+ for optical flow estimation, which achieves SOTA results on the DSEC dataset with the fastest inference speed. To further explore the impact of event density on model performance, we extend HREM to HREM+, a multi-density event dataset. Finally, we propose the Adaptive Density Module (ADM), which adjusts event density for improved flow estimation while maintaining fast inference speeds. Experiments demonstrate that ADM improves meshflow estimation by up to 8\% and optical flow estimation by up to 10\%.


%





\ifCLASSOPTIONcaptionsoff
  \newpage
\fi



%

{
  \bibliographystyle{IEEEtran}
  \bibliography{refs}
}
%
\vspace{-1.5cm}
\begin{IEEEbiography}[{\includegraphics[width=1in,height=1.25in,clip,keepaspectratio]{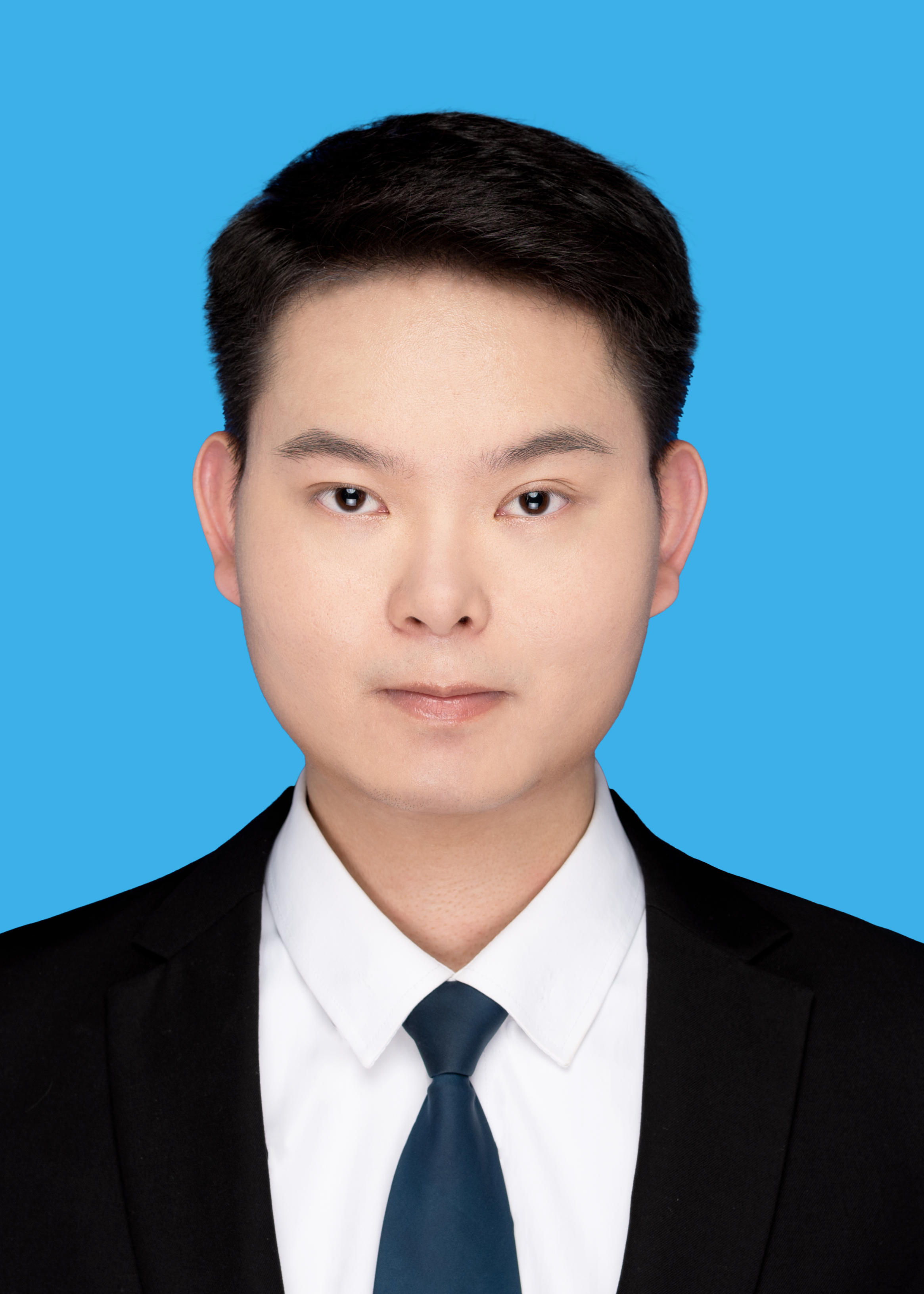}}]{Xinglong Luo}
received the B.S. degrees in the School of Electrical Engineering, Southwest Jiaotong University (SWJTU), in 2021 and M.S. degrees in the School of Information and Communication Engineering, University of Electronic Science and Technology of China (UESTC), Chengdu, China, in 2024. He is currently pursuing the Ph.D. degree with the School of Information and Communication Engineering, UESTC. His research interests include computer vision and event-based vision. 
\end{IEEEbiography}
\vspace{-1.5cm}
\begin{IEEEbiography}[{\includegraphics[width=1in,height=1.25in,clip,keepaspectratio]{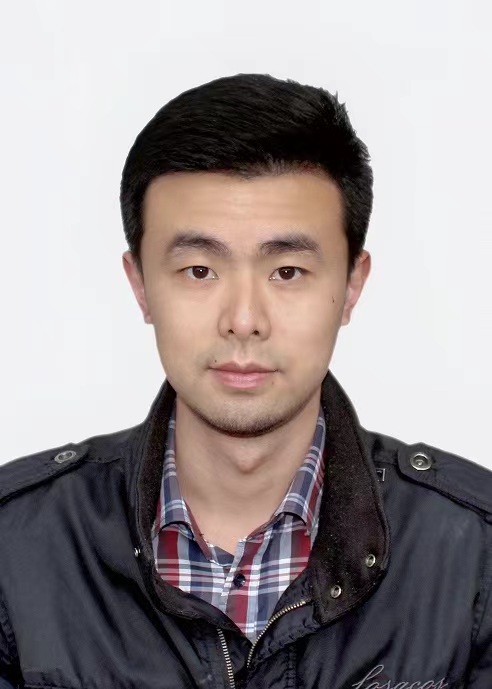}}]{Ao Luo}
 (Member, IEEE) received his B.E. degree from Southwest Jiaotong University (SWJTU), Chengdu, China, in 2013, and his Ph.D. degree from University of Electronic Science and Technology of China (UESTC), Chengdu, China, in 2020. He was a visiting scholar in State University of New York at Albany, and worked as a researcher at Megvii Research, Chengdu. Currently, he is an Assistant Professor in the School of Computing and Artificial Intelligence, SWJTU, Chengdu, China. His research interest focuses on deep learning for computer vision.
\end{IEEEbiography}

\begin{IEEEbiography}[{\includegraphics[width=1in,height=1.25in,clip,keepaspectratio]{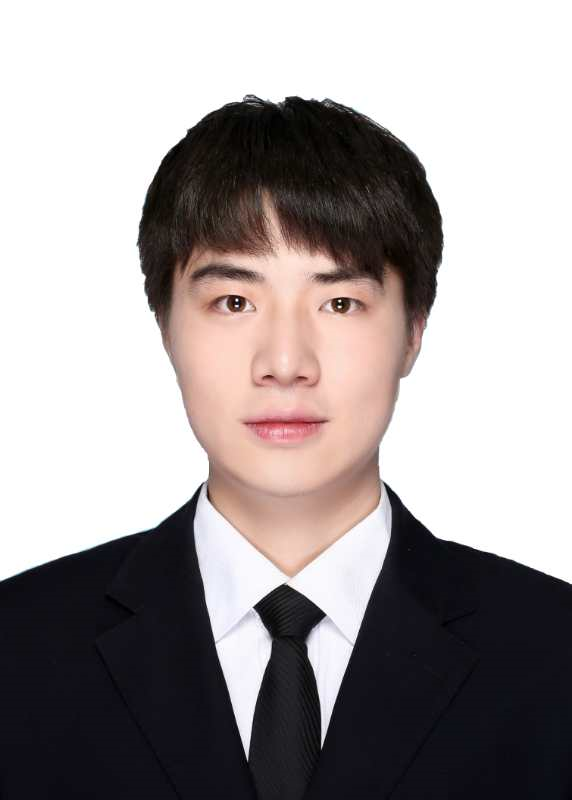}}]{Kunming Luo}
received his B.E. and M.S. degrees in the School of Information and Communication Engineering, the University of Electronic Science and Technology of China (UESTC), in 2016 and 2019. From 2019 to 2023, He worked as a research assistant of Megvii Technology, Chengdu. He is pursuing the Ph.D. degree in the Hong Kong University of Science and Technology (HKUST). His research interest includes computer vision and computer graphics.
\end{IEEEbiography}
\vspace{-1cm}
\begin{IEEEbiography}[{\includegraphics[width=1in,height=1.25in,clip,keepaspectratio]{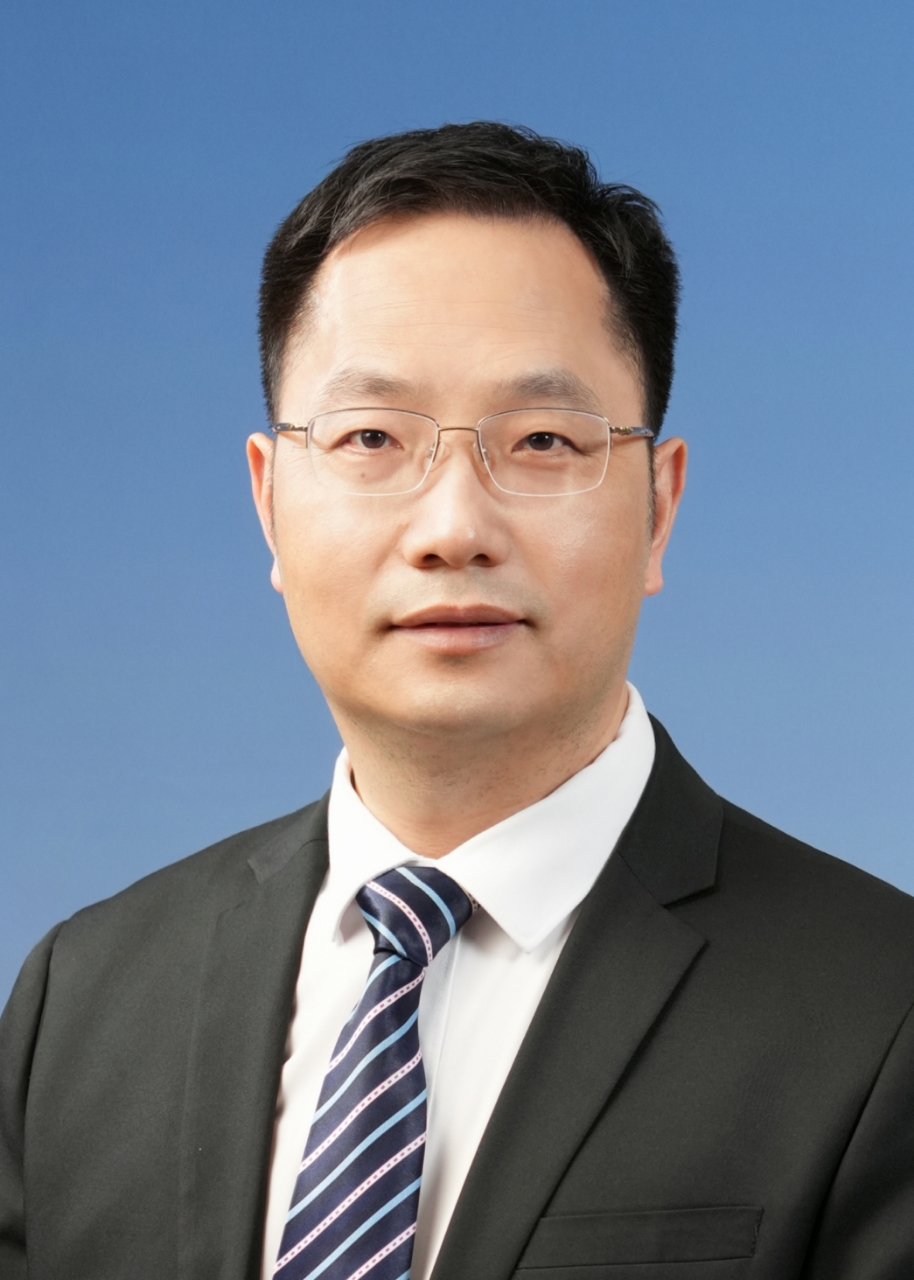}}]{Zhengning Wang}
 (Member, IEEE) received the B.E. and Ph.D. degrees from the Southwest Jiaotong University (SWJTU), Chengdu, China, in 2000 and 2007, respectively. From 2009 to 2011, he worked as a Post-Doctoral Fellow at the Second Research Institute of Civil Aviation Administration of China (CAAC), where he was the Project Leader of Remote Air Traffic Control Tower. In October 2014, he was visiting the Media Communication Lab, University of Southern California (USC), USA, as a Visiting Scholar for one year. He is currently a Professor at the School of Information and Communication Engineering, University of Electronic Science and Technology of China (UESTC). His research interests include image and video processing, computer vision, multimedia communication systems and applications, and intelligent transportation systems.
\end{IEEEbiography}
\vspace{-1cm}
\begin{IEEEbiography}[{\includegraphics[width=1in,height=1.25in,clip,keepaspectratio]{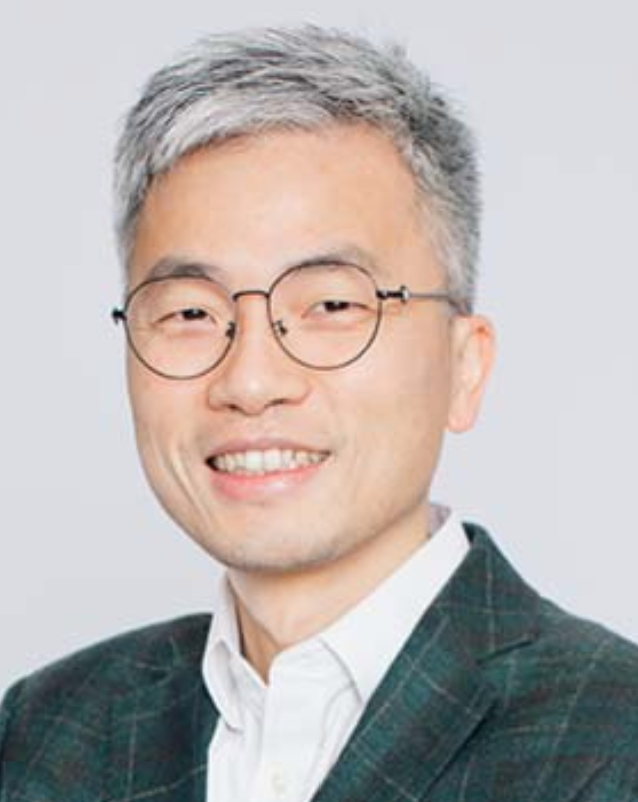}}]{Ping Tan}
(Senior Member, IEEE) is a Professor with the Department of Electronic and Computer Engineering, the Hong Kong University of Science and Technology (HKUST). He received the master’s and bachelor’s degrees from Shanghai Jiao Tong University (SJTU), China, in 2003 and 2000, respectively, and the PhD degree from the Hong Kong University of Science and Technology (HKUST) in 2007. He was an associate professor with the School of Computing Science, Simon Fraser University (SFU). Before that, he was an associate professor with the National University of Singapore (NUS). His research interests include computer vision, computer graphics, and robotics. He was an editorial board member of the IEEE Transactions on Pattern Analysis and Machine Intelligence, International Journal of Computer Vision, Computer Graphics Forum, and the Machine Vision and Applications, and was the area chairs for CVPR/ICCV, SIGGRAPH, SIGGRAPH Asia, and IROS.
\end{IEEEbiography}
\vspace{-1cm}
\begin{IEEEbiography}[{\includegraphics[width=1in,height=1.25in,clip,keepaspectratio]{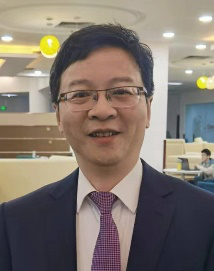}}]{Bing Zeng}
(M’91-SM’13-F’16) received the BEng and MEng degrees in electronic engineering from University of Electronic Science and Technology of China (UESTC), Chengdu, China, in 1983 and 1986, respectively, and the PhD degree in electrical engineering from Tampere University of Technology, Tampere, Finland, in 1991. He worked as a postdoctoral fellow at University of Toronto from September 1991 to July 1992 and as a Researcher at Concordia University from August 1992 to January 1993. He then joined the Hong Kong University of Science and Technology (HKUST). After 20 years of service at HKUST, he returned to UESTC inthe summer of 2013. At UESTC, he leads the Institute of lmage Processing to work on image and video processing, multimedia communicationcomputer vision, and Al technology.
\end{IEEEbiography}
\vspace{-1cm}
\begin{IEEEbiography}[{\includegraphics[width=1in,height=1.25in,clip,keepaspectratio]{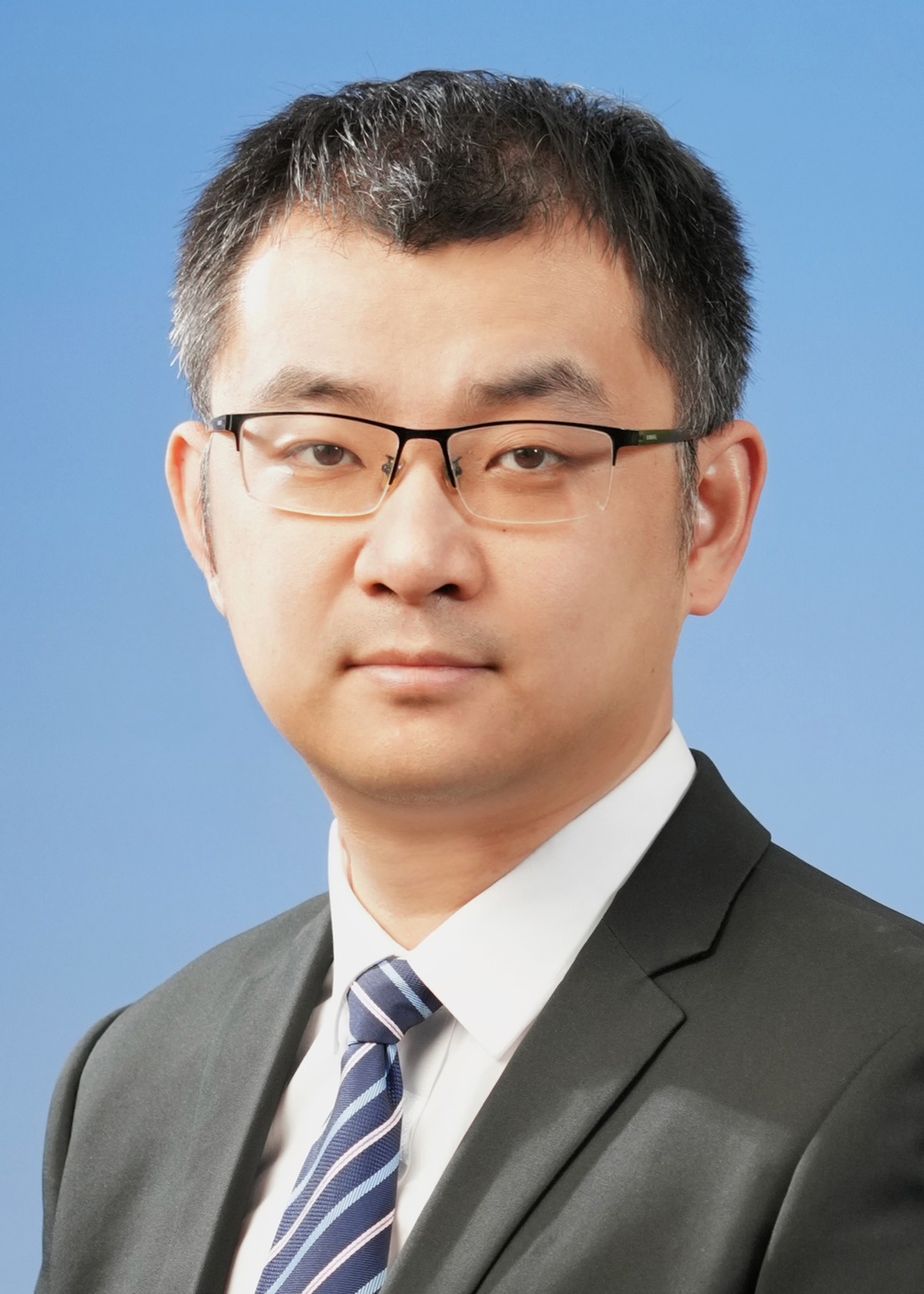}}]{Shuaicheng Liu} (M’14-SM’23) received the Ph.D. and M.Sc. degrees from the National University of Singapore, Singapore, in 2014 and 2010, respectively, and the B.E. degree from Sichuan University, Chengdu, China, in 2008. In 2015, he joined the University of Electronic Science and Technology of China (UESTC) andis currently a Professor with the Institute of lm-age Processing, School of Information and Com-munication Engineering, Chengdu, China. He works on computer vision, computer graphicsand computational imaging related problems, with applications in mobilephotography and videography.
\end{IEEEbiography}

\end{document}